\documentclass[10pt,twocolumn,letterpaper]{article}

\usepackage[pagenumbers]{cvpr}  

\usepackage{booktabs}
\usepackage{multirow}
\usepackage{graphicx}
\usepackage{caption}
\definecolor{lightblue}{RGB}{80, 160, 220}

\usepackage{currfile} 

\usepackage{caption} 

\definecolor{cvprblue}{rgb}{0.21,0.49,0.74}
\usepackage[pagebackref,breaklinks,colorlinks,allcolors=cvprblue]{hyperref}

\def\confName{CVPR}
\def\confYear{2027}

\title{FlexWAM: Learning Flexible Visual Routing  for World Action Models}

\author{
Mingyu Mei\textsuperscript{1,*} \quad
Haojie Xu\textsuperscript{1,*} \quad
Zibo Dai\textsuperscript{1} \quad
Hongjie Fang\textsuperscript{2} \quad
Shirun Tang\textsuperscript{3}
\\[0.6em]
Zhengrui Lv\textsuperscript{1} \quad
Yang Liu\textsuperscript{4} \quad
Guang Chen\textsuperscript{1,5} \quad
Xingyue Zhao\textsuperscript{1} \quad
Huiliang Shen\textsuperscript{1} \quad
Zaixing He\textsuperscript{1,\textdagger}
\\[0.8em]
{\normalsize
\textsuperscript{1} Zhejiang University
\quad
\textsuperscript{2} Shanghai Jiao Tong University
\quad
\textsuperscript{3} Noematrix
\quad
\textsuperscript{4} University of Minnesota
\quad
\textsuperscript{5} EndlessAI
}
\\[0.4em]
{\small
\textsuperscript{*}Equal contribution.
\qquad
\textsuperscript{\textdagger}Corresponding author.
}
}

\makeatletter

\newcommand{\Rmnum}[1]{%
    \expandafter\@slowromancap\romannumeral #1@}

\g@addto@macro\@maketitle{%
    \par
    \noindent
    \begin{minipage}{\textwidth}
        \centering
        \includegraphics[width=0.98\linewidth]{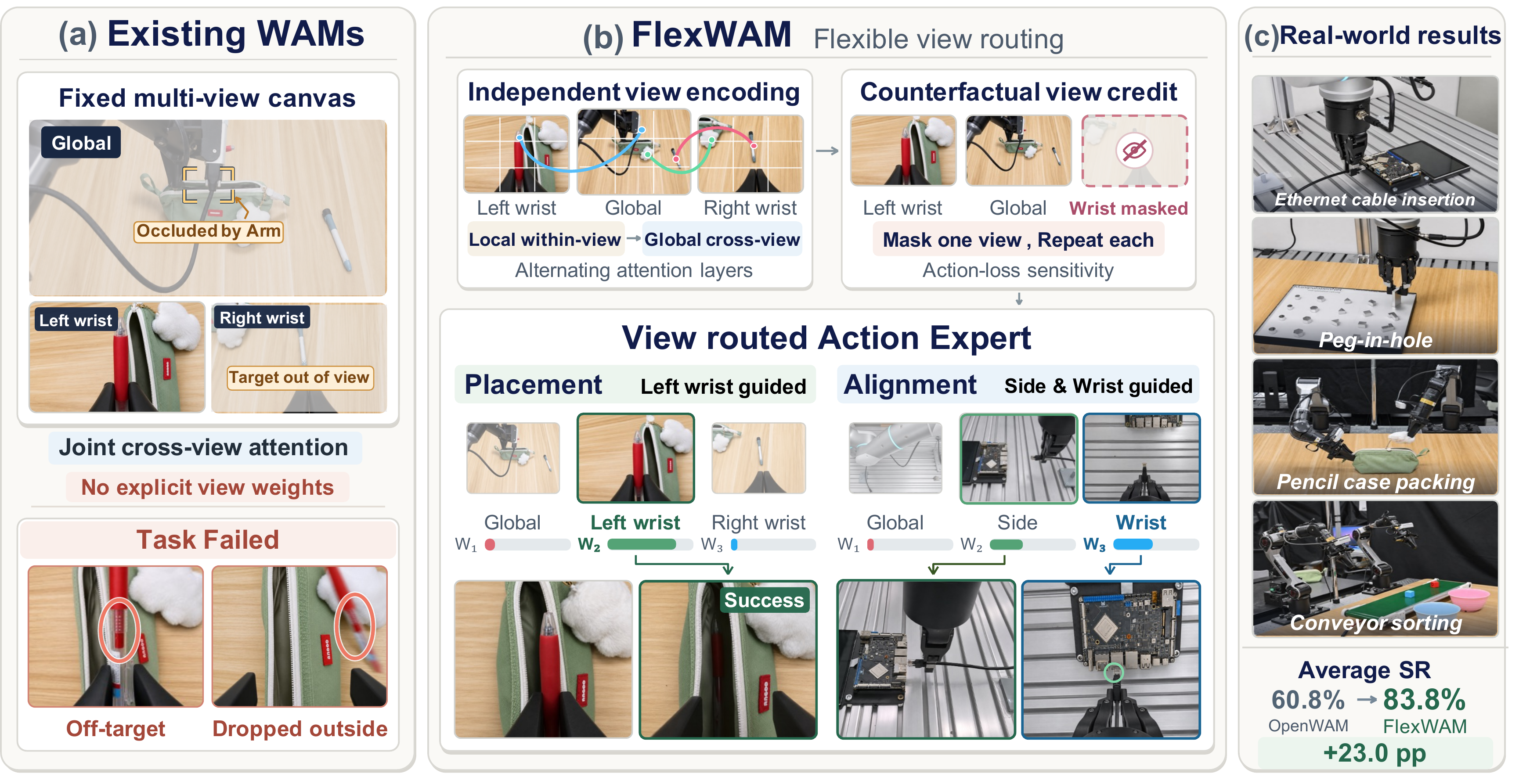}
        \captionof{figure}{Overview of FlexWAM. (a)  Existing WAMs fuse views in a fixed canvas and underutilize critical placement cues.  (b) FlexWAM combines within-view and cross-view world modeling and uses counterfactual masking to estimate each view’s contribution. The View-Routed Action Expert uses these estimates to adaptively weight the views for precise action prediction. (c) FlexWAM improves real-world performance across five challenging manipulation tasks on two robot platforms.}
        \label{fig:teaser}
        \vspace{1em}
    \end{minipage}
    \par
}

\makeatother

\begin{document}

\maketitle
\begin{abstract}
Humans dynamically adjust their reliance on sensory modalities across manipulation stages. Although existing World Action Models (WAMs) fuse multi-camera observations, they struggle to adapt their reliance on individual views based on the current observations. This raises a central question: how can the action expert integrate complementary visual evidence while prioritizing the views most relevant to the current action? FlexWAM addresses this challenge through three modules. First, we alternate within-view and cross-view attention to capture view-specific dynamics and complementary cues. For action prediction, View-Routed Action Attention uses an observation-conditioned router to prioritize relevant views. To calibrate routing, we train a critic using counterfactual view credit, estimated from action loss sensitivity upon suppressing each view’s contribution to action attention. We evaluate FlexWAM with two video backbones on five challenging real-world manipulation tasks across two robot platforms. The two variants improve mean task success by 16.4 and 23.0 percentage points over the strongest baseline. Project website: \color{lightblue}{https://flexwam.github.io/}.
\end{abstract}
\section{Introduction}
\label{sec:introduction}

Vision-Language-Action (VLA) models~\cite{pi0,pi05,openvla} and World Action Models (WAMs)~\cite{ye2026dreamzero,yuan2026fast,lingbotva} commonly use multi-camera observations for robotic manipulation. External cameras provide global workspace context, while wrist cameras capture local gripper--object relationships. In Fig.~\ref{fig:teaser}(a), the robot arm occludes the placement area in the external view, and the area lies outside the right wrist camera's field of view. Only the left wrist view reveals the local geometry needed for placement. Several WAMs tile camera images onto a fixed canvas, allocating fewer pixels and visual tokens to wrist views and making these fine details harder to preserve~\cite{ye2026dreamzero,motus,yuan2026fast,gigaworldpolicy05,wang2026openwam,flexpi}.

For video prediction, these models mix tokens from different cameras at every layer. When other views offer little relevant evidence, repeated mixing may interfere with modeling the local structure and motion in the left wrist view. \textbf{(1)Does predicting each view's future require cross-view interaction at every layer?} For action generation, standard action--video cross-attention assigns weights at the token level, leaving each camera's contribution implicit as observations and control demands change. \textbf{(2)How can the action expert integrate complementary information while prioritizing views most relevant to the current action?}

\textbf{Human perception dynamically prioritizes task-relevant sensory cues while integrating complementary evidence across modalities ~\cite{rohe2015cortical}.} At different manipulation stages, people adjust their reliance on sensory cues as control demands change~\cite{purdy1999manipulation,johansson2009coding}. When plugging in an Ethernet cable, for example, vision guides the approach and initial alignment of the connector. Once the connector touches the port, touch and force provide a more direct cue for detecting misalignment and making fine adjustments~\cite{park2026force}.

Inspired by this flexibility, we propose \textbf{FlexWAM} to address both questions(Fig.~\ref{fig:teaser}(b)) . \textbf{(1)} For video prediction, FlexWAM encodes each view independently and integrates complementary context through cross-view attention at alternating layers, with within-view attention modeling view-specific structure. \textbf{(2)} For action generation, an observation-conditioned router assigns per-view weights, and \textbf{View-Routed Action Attention} uses these weights to combine complementary cues while prioritizing the views most relevant to the current action.

When trained end to end with the action prediction loss alone, the action expert can improve by adapting to a nearly fixed mixture of views, leaving the routing weights close to their uniform initialization. To address this limitation, we introduce \textbf{counterfactual view credit}. We estimate each view's contribution by how much masking that view alone increases the action prediction loss. A larger increase indicates greater view importance. A \textbf{view critic} uses these credits to provide stable supervision for the router, enabling it to learn observation-dependent view weights without predefined task stages or view-importance annotations.

Our contributions are threefold:
\begin{itemize}
    \item We explore an \textbf{effective multi-view modeling} strategy alternating within-view and cross-view attention to capture view-specific dynamics and cross-view complementarity.

    \item We introduce \textbf{flexible view routing} to estimate view relative importance and \textbf{View-Routed Action Attention} to flexibly regulate visual attention during action generation.

    \item We propose \textbf{counterfactual view credit assignment} to estimate
    each view's relative contribution to action. A critic network uses these estimates to calibrate the router, mitigating shortcuts from action-loss-only training.
\end{itemize}

Experiments on five challenging real-world tasks using two robot platforms and two video-generation backbones demonstrate FlexWAM’s effectiveness. With Wan2.2-TI2V-5B~\cite{wan2025} and LingBot-Video-Dense (1.3B)~\cite{ma2026lingbotvideo,robbyant2026lingbotdense}, FlexWAM achieves mean task success rates of 77.2\% and 83.8\%, exceeding the strongest evaluated baseline by 16.4 and 23.0 percentage points, respectively.
\section{Related Work}
\label{sec:related_work}

\subsection{Vision-Language-Action Models}
Generalist robot policies learn language-conditioned manipulation across diverse tasks through large-scale multitask training~\cite{rt1,octo}.Vision-language-action (VLA) policies build on pretrained vision-language models to leverage visual and linguistic knowledge for robot action prediction~\cite{rt2,openvla}. For continuous action prediction, diffusion and flow-matching policies model conditional action distributions from multimodal observations~\cite{pi0,pi05,grootn1,rdt1b,smolvla,dexvla}. Beyond action modeling, recent work incorporates video prediction into policy to capture visual dynamics for robotic manipulation~\cite{vpp,gr1}. FlexWAM follows this direction by learning to use predictive visual representations from multiple cameras for action generation.

\subsection{World Action Models}
World action models (WAMs) combine visual dynamics prediction with action learning. Prior work recovers actions from generated visual trajectories~\cite{unipi,vidar,robodreamer}, conditions action decoders on predictive video features~\cite{vpp,mimicvideo,dit4dit}, or jointly models video and actions~\cite{gr1,uwm,uva,ye2026dreamzero,cosmospolicy}. Other studies explore causal temporal modeling~\cite{lingbotva} and latent action representations~\cite{motus}. To reduce inference cost, Fast-WAM~\cite{yuan2026fast} and related methods~\cite{gigaworldpolicy05,dit4dit} retain future-video prediction as training supervision and predict actions without completing future-video generation at deployment. Multi-camera WAMs often combine views in a fixed spatial layout by concatenating images~\cite{ye2026dreamzero,motus,yuan2026fast,gigaworldpolicy05,wang2026openwam,flexpi} or latent features~\cite{lingbotva,tau0wm}. Other methods maintain separate latent streams and exchange information through cross-view attention~\cite{genieenvisioner,wallwm,geact2}. For video prediction, FlexWAM encodes each view separately and alternates within-view and cross-view attention to capture view-specific dynamics and cross-view complementarity.

\subsection{Adaptive Visual Perception for Robot Control}

The relevance of each camera view to action prediction varies across manipulation stages. Active perception methods adjust viewpoints or observation regions~\cite{tvve,activevla,maeselect}, and visual token selection aims to retain task-relevant tokens from available observations~\cite{lightvla,vlapruner,specprunevla,efficientvla}. Prior work also adapts sensory inputs through stage-dependent masking~\cite{longvla}, multisensory weighting~\cite{playtothescore}, or target-region highlighting in selected views~\cite{hint}. BFA~\cite{bfa} and Selective Perception~\cite{selectiveperception} learn camera or modality weights from VLM-assisted importance annotations, while FocusVLA~\cite{focusvla} guides view routing with future interaction intent learned from rule-based trajectory labels. For action prediction, FlexWAM’s Action Expert uses an observation-conditioned router to dynamically adjust its reliance on each camera view. To calibrate the router, FlexWAM estimates counterfactual view credit by measuring changes in action loss when each view branch is masked, without additional task-stage or view-importance annotations.


\section{Method}
\label{sec:method}

\begin{figure*}[t]
    \centering
    \includegraphics[width=1\linewidth]{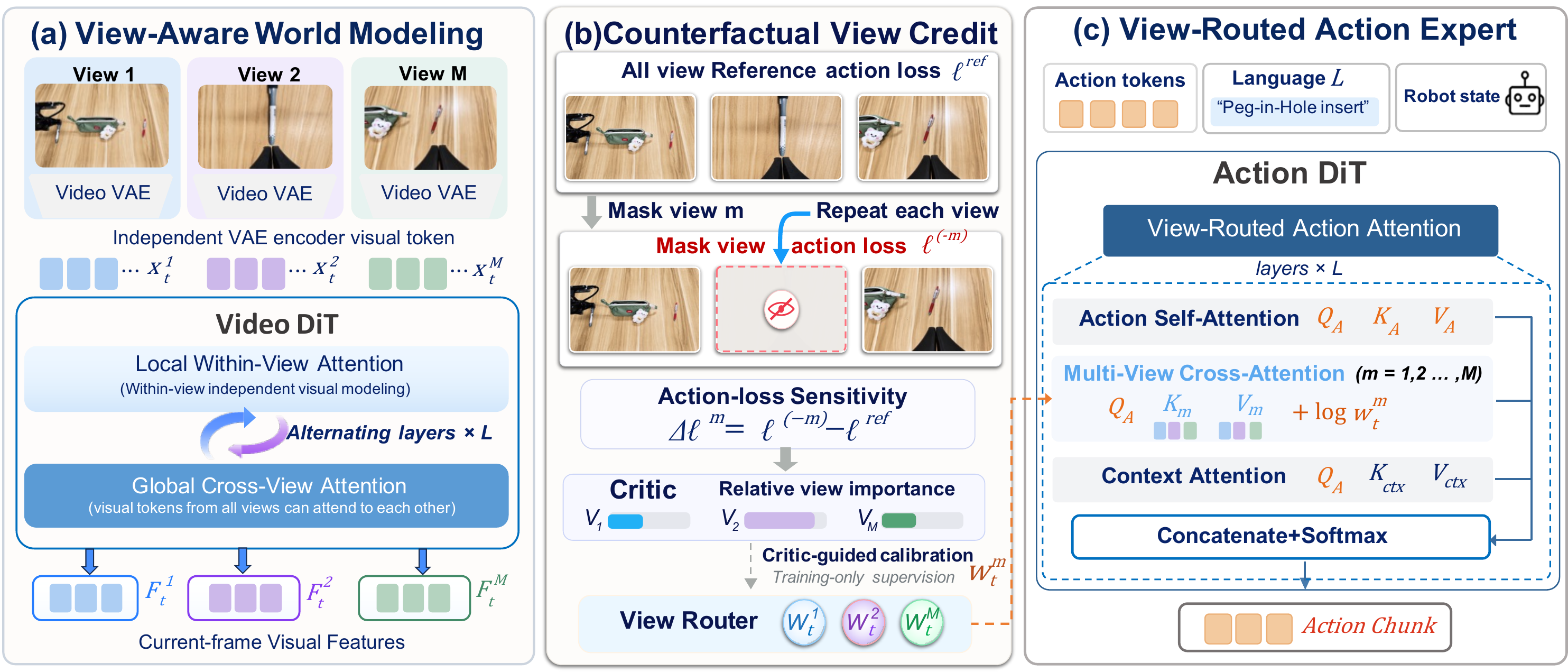}
    \caption{Framework of FlexWAM. (a) Alternating within-view and cross-view attention captures view-specific dynamics and integrates cross-view context. (b) The critic calibrates the router through counterfactual evaluations during training. (c) View-Routed Action DiT flexibly uses multi-view routing for action prediction. }
    \label{fig:framework}
    \vspace{-1em}
\end{figure*}

\subsection{Problem Formulation}
\label{sec:problem_formulation}

\noindent\textbf{WAM formulation.}
At control step $t$, the observation $o_t=(\{I_t^m\}_{m=1}^M,\mathbf s_t)$
comprises $M$ camera images and the robot's proprioceptive state.
Given $o_t$ and a language instruction $\ell$, FlexWAM models the
conditional distribution of an $H$-step action chunk
$\mathbf A_t=(\mathbf a_t,\ldots,\mathbf a_{t+H-1})$:
\begin{equation}
\label{eq:policy_formulation}
p_{\theta,\rho}(\mathbf A_t \mid o_t,\ell).
\end{equation}
Here, $\theta=(\theta_v,\theta_a)$ comprises the parameters of the
video and action experts, respectively, and $\rho$ denotes the
parameters of the view router. Following Fast-WAM~\cite{yuan2026fast}, we jointly train a video DiT
to predict future visual latents and an action expert to generate
action chunks. At inference, we skip future latent prediction and
condition the action expert on visual features from the current
observation.

Both experts use flow matching. For target $\mathbf y$, Gaussian noise
$\boldsymbol\epsilon\sim\mathcal N(\mathbf 0,\mathbf I)$, and noise level $\sigma$, the
input is $\mathbf y_\sigma=(1-\sigma)\mathbf y+\sigma\boldsymbol\epsilon$
and the target velocity is $\boldsymbol\epsilon-\mathbf y$.
The squared velocity error $\mathcal L_{\mathrm{FM}}$ gives
$\mathcal L_{\mathrm{vid}}$ for future visual latents and
$\mathcal L_{\mathrm{act}}$ for action chunks.

\noindent\textbf{Key Challenges.} Existing WAMs often spatially concatenate multi-camera observations
before video encoding, treating heterogeneous views as a unified
representation~\cite{ye2026dreamzero,motus,yuan2026fast,gigaworldpolicy05,wang2026openwam,flexpi}.
This design presents three limitations:
\textbf{(i)} fixed spatial budgets may reduce the resolution of views
that are critical at particular manipulation stages;
\textbf{(ii)} joint attention and denoising may introduce unnecessary
interactions between views with distinct visual distributions and dynamics;
and \textbf{(iii)} fixed spatial layouts complicate adaptation to heterogeneous camera configurations across pretraining datasets and post-training finetune tasks. 
\textbf{The key challenges are to balance view-specific dynamics modeling with cross-view fusion and to estimate each view's
observation-dependent contribution to action prediction.}
FlexWAM addresses these challenges through three components:
view-aware world modeling to structure video attention within
and across views, flexible view routing to allocate attention across cameras, and counterfactual view credit assignment to calibrate routing.

\subsection{View-Aware World Modeling}
\label{sec:view_modeling}

\noindent\textbf{Local Within-View Attention.}
Multi-view world modeling requires coherent dynamics within each
camera stream and information exchange across cameras. As shown in Fig.~\ref{fig:framework}(a),
each camera stream is independently encoded by a frozen video VAE,
preserving its spatial token grid. In our real-world experiments, the video VAE receives images at a resolution of $192 \times 224$ pixels(height $\times$ width).
Camera embeddings distinguish the streams, while all views share
the same video DiT parameters and language conditioning.

We follow the temporal visibility rules of
Fast-WAM~\cite{yuan2026fast} and introduce alternating
within-view and cross-view self-attention, inspired by the
alternating-attention design of VGGT~\cite{wang2025vggt}.
Within-view layers apply shared attention independently to
each camera stream, modeling its appearance and temporal dynamics.
Visual tokens attend to temporally visible tokens from the
same camera.

\noindent\textbf{Global Cross-View Attention.}
Global cross-view layers integrate complementary information
by allowing visual tokens to attend to temporally visible
tokens across all cameras.
Both attention patterns operate under the same temporal
visibility constraints.
Let $\mathbf X^m$ denote all spatiotemporal tokens from view $m$
at the current layer, and let
$\mathbf X = [\mathbf X^1;\ldots;\mathbf X^M]$
be their concatenation.
For each attention head, the two attention operations are
\begin{equation}
\label{eq:view_attention}
\begin{aligned}
\operatorname{Attn}_{\mathrm{local}}(\mathbf X^m)
&=
\operatorname{softmax}\left(
\frac{\mathbf Q^m(\mathbf K^m)^{\top}}{\sqrt{d_h}}
\right)\mathbf V^m,
\\
\operatorname{Attn}_{\mathrm{global}}(\mathbf X)
&=
\operatorname{softmax}\left(
\frac{\mathbf Q\mathbf K^{\top}}{\sqrt{d_h}}
\right)\mathbf V,
\end{aligned}
\end{equation}
where $\mathbf Q^m$, $\mathbf K^m$, and $\mathbf V^m$
are the query, key, and value projections of $\mathbf X^m$.
The global queries are
$\mathbf Q = [\mathbf Q^1;\ldots;\mathbf Q^M]$,
with $\mathbf K$ and $\mathbf V$ defined analogously.
Here, $d_h$ is the head dimension, and the projection parameters
are specific to each layer.
Softmax normalization is restricted to keys allowed by the
temporal visibility constraints.

\noindent\textbf{View-Balanced Supervision.}
Future latents $\mathbf Z^m$ are predicted for each view, with
$\mathcal{L}_{\mathrm{vid}} =
\frac{1}{M}\sum_{m=1}^{M}
\mathcal{L}_{\mathrm{FM}}(\mathbf Z^m)$.
The loss is averaged over valid future tokens within each view
and then equally across views.

\subsection{Flexible View Routing for Action Modeling}
\label{sec:view_routing}

\noindent\textbf{View Weight Prediction.}
The relevance of each camera stream to action prediction varies
with the current observation.
We introduce a view router that predicts observation-dependent
view weights.
Given features
$\mathcal F_t=\{\mathbf F_t^m\}_{m=1}^M$
extracted by the video DiT from the clean observation frame, we construct a compact summary $\mathbf h_t^m$
for each view using two parallel branches:
coarse spatial pooling and learned-query attention over
its spatial tokens.
Their outputs are concatenated to form each view summary.
The feature-extraction layer and summary configuration are
specified in the supplementary material.
The concatenated view summaries are fed into a lightweight
MLP $R_{\mathrm{mlp}}$ to predict normalized routing weights:
\begin{equation}
\label{eq:router}
\mathbf r_t
=
\operatorname{softmax}\left(
R_{\mathrm{mlp}}([\mathbf h_t^1;\ldots;\mathbf h_t^M])
\right),
\qquad
w_t^m = M r_t^m.
\end{equation}

The router parameters $\rho$ comprise the trainable parameters of the view-summary modules and $R_{\mathrm{mlp}}$. Visual features are detached before summarization, preventing gradients from propagating into the video DiT. Zero-initializing the final MLP projection yields $r_t^m=1/M$ and $w_t^m=1$, resulting in zero routing bias at initialization.

\noindent\textbf{View-Routed Action Attention.}
We inject the predicted view weights as additive biases
to visual attention logits.
For each head at layer $k$, the action expert provides
$\mathbf Q_{\mathrm A}^k,\mathbf K_{\mathrm A}^k,
\mathbf V_{\mathrm A}^k\in\mathbb R^{H\times d_h}$,
where the subscript $\mathrm A$ denotes action tokens.
The video expert provides
$\mathbf K_m^k,\mathbf V_m^k\in\mathbb R^{N_m\times d_h}$
for the $N_m$ current-frame spatial tokens of camera $m$.
The subscript $\mathrm{ctx}$ denotes context from robot state
and language, with
$\mathbf K_{\mathrm{ctx}}^k,\mathbf V_{\mathrm{ctx}}^k$
collecting the context keys and values participating
in this attention.

As shown in Fig.~\ref{fig:framework}(c), the action-to-action, context, and action-to-view
score matrices are:
\begin{equation}
\label{eq:routed_attention_scores}
\begin{aligned}
\mathbf S_{\mathrm A}^k
&=
\frac{\mathbf Q_{\mathrm A}^k(\mathbf K_{\mathrm A}^k)^\top}{\sqrt{d_h}},
\qquad
\mathbf S_{\mathrm{ctx}}^k
=
\frac{\mathbf Q_{\mathrm A}^k(\mathbf K_{\mathrm{ctx}}^k)^\top}{\sqrt{d_h}},
\\
\mathbf S_m^k
&=
\frac{\mathbf Q_{\mathrm A}^k(\mathbf K_m^k)^\top}{\sqrt{d_h}}
+
\log w_t^m,
\end{aligned}
\end{equation}
where each scalar $\log w_t^m$ is broadcast across the
corresponding view's score matrix.
The score matrices are concatenated along the key dimension
and normalized jointly:
\begin{equation}
\label{eq:routed_attention}
\mathbf P^k
=
\operatorname{softmax}\left(
[\mathbf S_{\mathrm A}^k,\mathbf S_1^k,\ldots,
\mathbf S_M^k,\mathbf S_{\mathrm{ctx}}^k]
\right),
\end{equation}
where softmax is applied row-wise over all visible keys.
Partitioning $\mathbf P^k$ according to these key groups
gives the attention output:
\begin{equation}
\label{eq:routed_attention_output}
\begin{aligned}
\mathbf O_{\mathrm A}^k
&=
\mathbf P_{\mathrm A}^k\mathbf V_{\mathrm A}^k
+
\sum_{m=1}^{M}\mathbf P_m^k\mathbf V_m^k
+
\mathbf P_{\mathrm{ctx}}^k\mathbf V_{\mathrm{ctx}}^k.
\end{aligned}
\end{equation}
The first term captures bidirectional self-attention within
the action chunk.
The routing design provides two benefits:
\textbf{(i)} adding $\log w_t^m$ introduces an
observation-dependent view prior, adapting each camera's
contribution while preserving relative spatial token
preferences within each view;
\textbf{(ii)} by restricting this prior to visual logits,
it further preserves the scoring mechanisms of
action self-attention and context conditioning, while the
shared softmax coordinates their contributions with the
routed visual information.

\noindent\textbf{Action-Driven Learning.}
Gradients from the action loss reach the router through the attention
biases, and the view gates are shared across action layers. During joint
training, the action expert can reduce the prediction loss by adapting its own parameters to nearly uniform routing weights, while the router's weights remain largely unchanged across observations. This motivates a separate supervision signal that evaluates each view's contribution to action prediction.

\subsection{Counterfactual View Credit Assignment}
\label{sec:counterfactual_credit}

\noindent\textbf{Counterfactual View Evaluation.} As shown in Fig.~\ref{fig:framework}(b), 
we therefore introduce \textbf{counterfactual(CF) view evaluation}
to estimate each view's relative importance and calibrate the router.
We hold the model parameters fixed and measure the change in action
loss when each visual branch is masked in action attention.

To prevent the counterfactual evaluation from amplifying existing biases in the routing weights, we set all reference
gates to one. To remove camera $m$, we set its gate in
Eq.~\eqref{eq:routed_attention_scores} to zero and keep all other
gates at one.

\begin{equation}
\label{eq:view_intervention}
w_j^{\mathrm{ref}}=1,
\qquad
w_j^{(-m)}
=
\begin{cases}
0, & j=m,\\
1, & j\ne m,
\end{cases}
\quad j=1,\ldots,M.
\end{equation}
For each training example, we hold the model parameters fixed and use the same sampled action-noise level and Gaussian noise realization for the all-view reference and all masked forward passes. The per-example loss change for masking branch \(m\) is:
\begin{equation}
\label{eq:counterfactual_delta}
\Delta\ell_{\mathrm{act},t}^{m}
=
\ell_{\mathrm{act},t}(\mathbf w^{(-m)})
-
\ell_{\mathrm{act},t}(\mathbf w^{\mathrm{ref}}).
\end{equation}
A larger loss increase indicates greater reliance on that branch given
the remaining inputs. We standardize the signed changes within each
observation to keep the target scale stable.
\begin{equation}
\label{eq:counterfactual_normalization}
z_t^m
=
\frac{
\Delta\ell_{\mathrm{act},t}^{m}
-
\mu_{\Delta\ell,t}
}{
s_{\Delta\ell,t}+\varepsilon
},
\end{equation}
where $\mu_{\Delta\ell,t}$ and $s_{\Delta\ell,t}$ are the across-view
mean and standard deviation, with $\varepsilon>0$ for numerical stability.

\noindent\textbf{Critic-Guided Router Calibration.}
Counterfactual (CF) scores depend on the current action predictor and can drift as it is optimized. Even for a fixed predictor and training example, different action-noise levels can change the relative loss increases across views. Directly converting these scores into a routing distribution therefore exposes the router to variation across individual CF evaluations.

We train a separate critic $C_\phi$, with its own parameters
$\phi$, to predict one normalized CF score per view from
detached current-frame visual features. It uses the router
architecture.
\begin{equation}
\label{eq:critic_prediction}
\widehat{\mathbf z}_t
=
C_\phi(\operatorname{sg}(\mathcal F_t)).
\end{equation}
where $\operatorname{sg}$ denotes stop-gradient.
The critic regression loss $\mathcal L_{\mathrm{critic}}$
is the mean squared error between its predictions and the
periodically refreshed, detached CF scores.
Regression across sampled perturbations encourages the critic
to approximate the expected normalized CF scores conditioned
only on visual features. We convert its predictions into a
detached soft target for router calibration.

\begin{equation}
\label{eq:route_kl}
\begin{aligned}
\mathbf r_t^{\mathrm{teach}}
&=
\operatorname{softmax}
\left(\widehat{\mathbf z}_t/\tau\right),\\
\mathcal L_{\mathrm{route}}
&=
D_{\mathrm{KL}}
\left(
\operatorname{sg}(\mathbf r_t^{\mathrm{teach}})
\,\|\,\mathbf r_t
\right).
\end{aligned}
\end{equation}
Here $\tau>0$ controls the sharpness of the target distribution. The critic regression loss updates only the critic, and the calibration loss updates only the router. The action expert receives the router’s predicted weights through the visual attention biases.

\noindent\textbf{Training and Inference.}
Training proceeds in four stages. We first train the video and action experts with uniform view gates, then train the router with the action loss. In the third stage, we fit the critic to CF scores while the router continues to learn from the action loss. In the fourth stage, we add critic-guided KL calibration to the router objective and continue training the critic. Once calibration begins, we linearly warm up the KL weight \(\beta\) from 0.33 to 0.50. The objectives for the active parameter groups are:
\begin{equation}
\label{eq:parameter_objectives}
\begin{aligned}
\mathcal J_{\theta}
&= \mathcal L_{\mathrm{act}} + \lambda_v\mathcal L_{\mathrm{vid}},\\
\mathcal J_{\rho}
&= \mathcal L_{\mathrm{act}} + \beta\mathcal L_{\mathrm{route}},\\
\mathcal J_{\phi}
&= \mathcal L_{\mathrm{critic}}.
\end{aligned}
\end{equation}
At inference, the router computes the view weights once per observation
and reuses them across all action-denoising steps. Counterfactual
evaluations and the critic are used only during training.

\section{Experiments}

We design our experiments to answer four questions:

\begin{figure*}[t]
    \centering
    \includegraphics[page=1,width=0.98\textwidth]{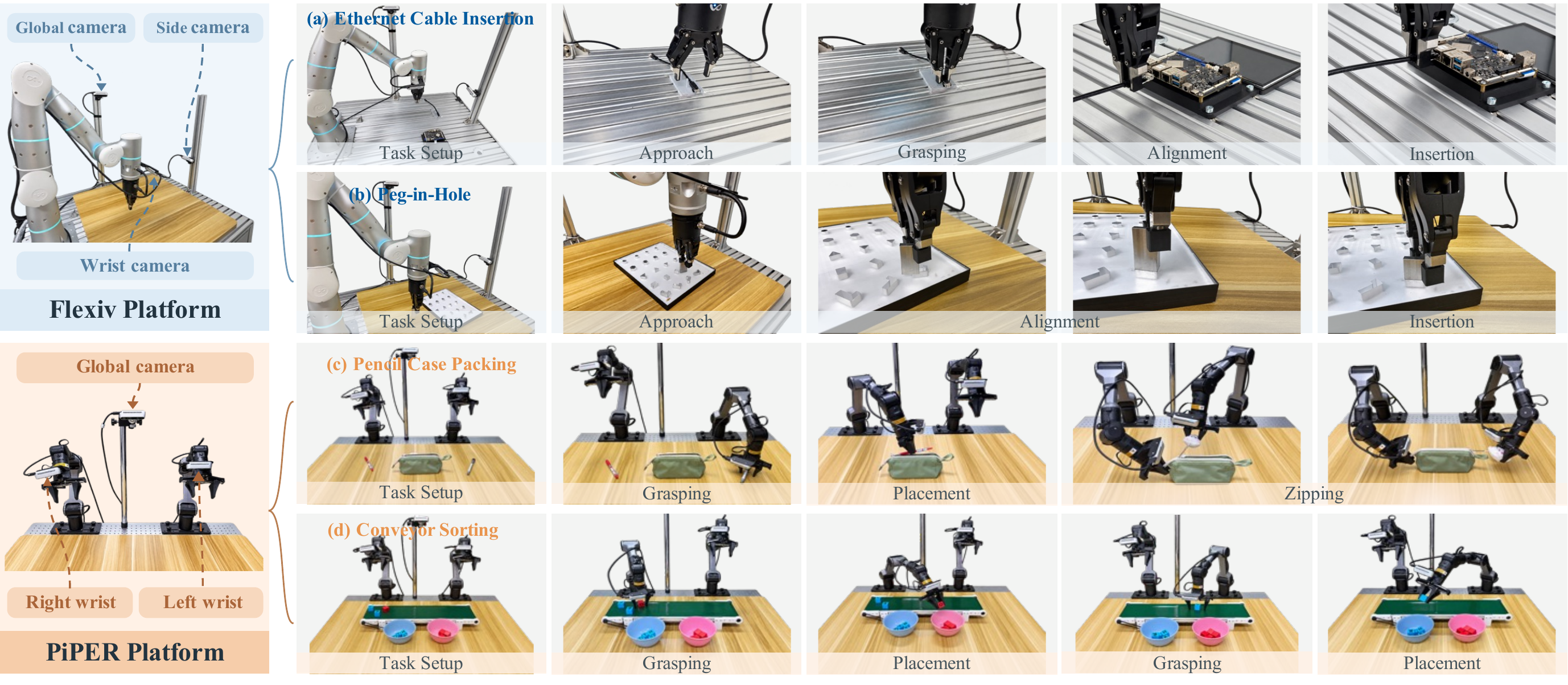}
    \caption{Real-world robot platforms, camera placements, and task stages.
    The single-arm Flexiv performs (a) Ethernet cable insertion and (b) peg-in-hole insertion.
    The dual-arm PiPER performs (c) pencil case packing and (d) conveyor sorting.}
    \label{fig:experimental_setup}
    \vspace{-0.5em}
\end{figure*}

\begin{table*}[t]
    \centering
    \caption{Real-world task success rates (\%).
    Pens Placement is the mean placement success rate across the two pens.
    Pencil Case Overall requires successful packing and zipper closure.
    Average is computed over the five tasks, using Pencil Case Overall for pencil case packing. The highest value in each column is bolded, and the second-highest distinct value is underlined. }
    \label{tab:real_world_results}
    \footnotesize
    \setlength{\tabcolsep}{3pt}
    \renewcommand{\arraystretch}{1.05}
    \begin{tabular*}{\textwidth}{@{\extracolsep{\fill}}lccccccc@{}}
        \toprule
        \multirow{2}{*}{Method}
        & \multicolumn{3}{c}{\textbf{Dual-Arm PiPER}}
        & \multicolumn{3}{c}{\textbf{Single-Arm Flexiv}}
        & \multirow{2}{*}{\textbf{Average}} \\
        \cmidrule(lr){2-4}
        \cmidrule(lr){5-7}
        & \textbf{\textit{Pens Placement}}
        & \textbf{\textit{Pencil Case Overall}}
        & \textbf{\textit{Conveyor Sorting}}
        & \textbf{\textit{Peg-in-Hole (L)}}
        & \textbf{\textit{Peg-in-Hole (Round)}}
        & \textbf{\textit{Cable Insertion}}
        & \\
        \midrule
        Fast-WAM~\cite{yuan2026fast}
        & 60.0\% & 40.0\% & 80.0\%
        & 60.0\% & 55.0\% & 55.0\% & 58.0\% \\
        OpenWAM~\cite{wang2026openwam}
        & 67.5\% & 55.0\% & \textbf{89.0\%}
        & \underline{70.0\%} & 50.0\% & 40.0\% & 60.8\% \\
        LaWAM~\cite{chen2026lawam}
        & 67.5\% & 40.0\% & 67.0\%
        & 5.0\% & 20.0\% & 5.0\% & 27.4\% \\
        $\pi_{0.5}$~\cite{pi05}
        & 62.5\% & 55.0\% & \underline{88.0\%}
        & 50.0\% & 50.0\% & 35.0\% & 55.6\% \\
        \midrule
        FlexWAM (Wan)
        & \textbf{80.0\%} & \textbf{65.0\%} & 86.0\%
        & \textbf{95.0\%} & \underline{80.0\%} & \underline{60.0\%}
        & \underline{77.2\%} \\
        FlexWAM (LingBot)
        & \underline{72.5\%} & \underline{60.0\%} & \textbf{89.0\%}
        & \textbf{95.0\%} & \textbf{100.0\%} & \textbf{75.0\%}
        & \textbf{83.8\%} \\
        \bottomrule
    \end{tabular*}
    \vspace{-1em}
\end{table*}

\textbf{(Q1)}
Does FlexWAM consistently improve performance across tasks and robot platforms?

\textbf{(Q2)}
How do view routing and counterfactual calibration affect action precision at critical manipulation stages?

\textbf{(Q3)}
How should view-specific information and cross-view interaction be balanced?

\textbf{(Q4)}
How do individual components affect performance and view contribution learning?

\subsection{Experimental Setup}
\noindent\textbf{Platforms and Tasks.} We conduct real-world experiments on a dual-arm AgileX PiPER and a single-arm Flexiv Rizon 4s.
The PiPER setup uses one global camera and two wrist-mounted Intel RealSense D435 RGB-D cameras.
The Flexiv setup uses two fixed cameras and one wrist-mounted Intel RealSense D415 RGB-D camera.
We evaluate five tasks across the two platforms (Fig.~\ref{fig:experimental_setup}).
PiPER performs pencil case packing and conveyor sorting.
Flexiv performs L-shaped peg insertion, round peg insertion, and Ethernet cable insertion.
Both peg-in-hole tasks use the most challenging clearance (0.02~mm) in ManipulationNet~\cite{chen2026manipulationnet}.
We collect 100 teleoperated demonstrations per task at 30~Hz, recording multi-view images and proprioceptive states.

\noindent\textbf{Baselines.}
We compare FlexWAM with three representative WAMs: Fast-WAM~\cite{yuan2026fast}, OpenWAM~\cite{wang2026openwam}, and LaWAM~\cite{chen2026lawam} and  the state-of-the-art VLA model $\pi_{0.5}$~\cite{pi05}. We evaluate FlexWAM with two video-generation backbones: Wan2.2-TI2V-5B~\cite{wan2025} and LingBot-Video-Dense (1.3B)~\cite{ma2026lingbotvideo,robbyant2026lingbotdense}.
We denote FlexWAM with the Wan and LingBot backbones as 
\emph{FlexWAM(Wan)} and \emph{FlexWAM(LingBot)}, respectively, 
throughout the paper.

\noindent\textbf{Metrics and Evaluation Protocol.}
We use Success Rate (SR) as the primary metric. To evaluate performance and inference latency, models are deployed locally (RTX 3090 for PiPER, RTX 4090 for Flexiv) and remotely (H200 for both). We conduct 20 trials per static task, and 100 blocks for conveyor sorting. To ensure strict fairness, all baselines and ablations are evaluated on the \textit{exact same} 20 randomized positions for static tasks, while the dynamic \textit{conveyor} task follows an identical placement protocol across all methods.

\subsection{Results}

\begin{figure*}[t]
    \centering
    \includegraphics[page=1,width=1\textwidth]{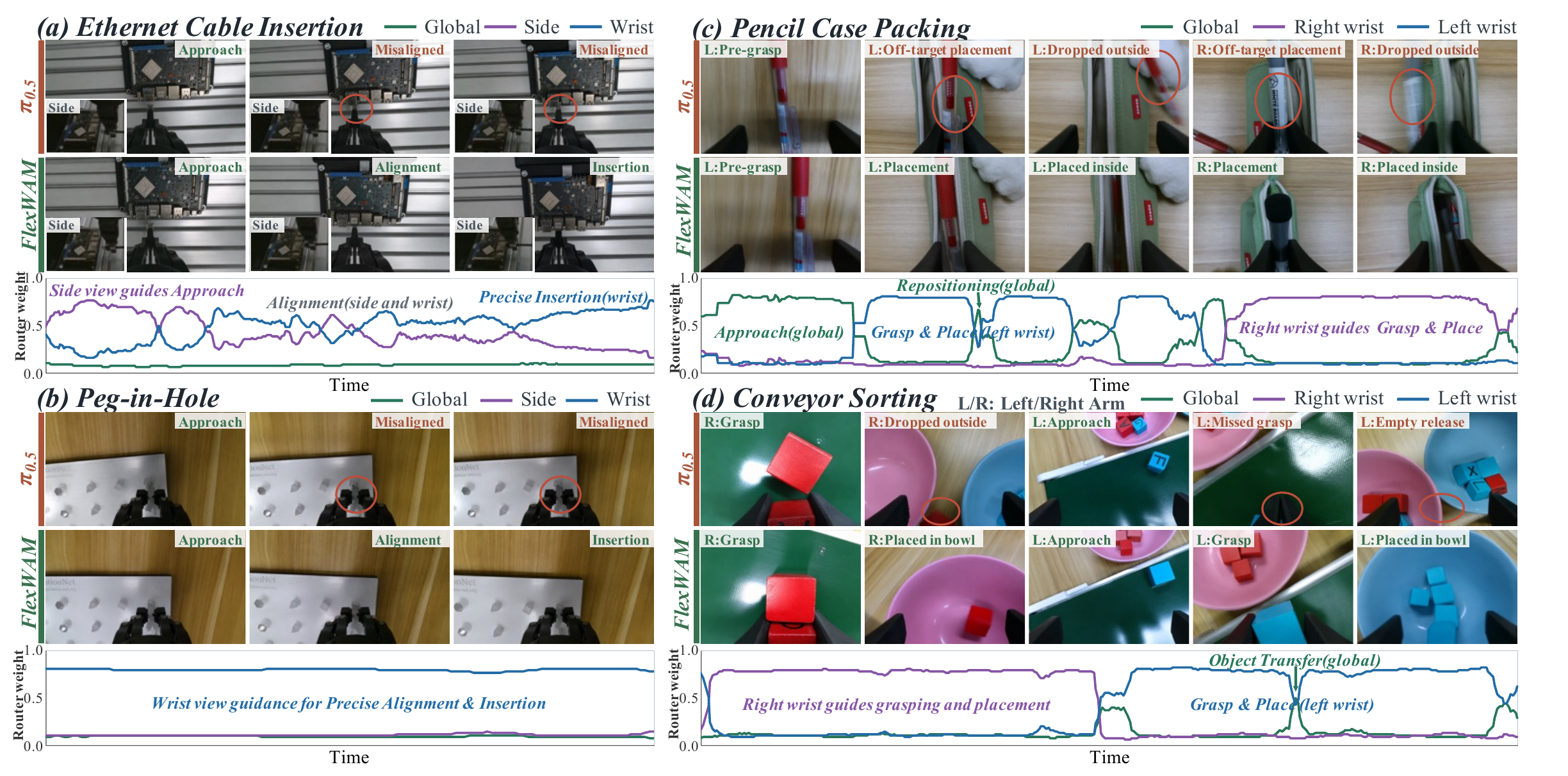}
    \caption{FlexWAM's adaptive camera routing at key manipulation stages. Each panel contrasts representative $\pi_{0.5}$~\cite{pi05} failure modes with how FlexWAM (Wan) adapts its actions at the same stages. The curves show how FlexWAM's routing weights for the three camera views change over time. L and R denote the wrist-camera views of the left and right arms, respectively.}
    \label{fig:router}
    \vspace{-1em}
\end{figure*}

 \textbf{FlexWAM demonstrates robust performance across diverse manipulation settings, covering extremely precision, high dynamics, and deformable object manipulation (Q1)}. As shown in Table~\ref{tab:real_world_results}, FlexWAM average success rates reach 77.2\% with Wan and 83.8\% with LingBot, outperforming all compared methods and supporting its effectiveness across the evaluated camera configurations, robot platforms, and video foundation models.
 
 \textbf{The clearest gains occur in precision tasks Peg-in-Hole and Cable Insertion.} During fine alignment, robot self-occlusion limits fixed-camera visibility, whereas the wrist camera provides a clearer view of the relative object–target configuration. Despite robot-data pretraining, $\pi_{0.5}$~\cite{pi05} and OpenWAM~\cite{wang2026openwam} sometimes block or repeatedly attempt insertion near the target without resolving residual misalignment. In contrast, FlexWAM's CF-calibrated router, as shown in Fig.~\ref{fig:router}(a,b), increases the contribution of the wrist-view branch during fine alignment, enabling the policy to make precise corrections for successful insertion.

  In dual-arm tasks, \textbf{FlexWAM flexibly adjusts the contribution of three views according to the manipulation stage.} In pencil case packing task, it prioritizes the active arm’s wrist view during grasping and placement (Fig.~\ref{fig:router}(c)), improving precision when placing pens through the narrow opening. Conveyor sorting may require relatively less precision, and the global view could already provide sufficient information for task completion, potentially reducing the need for dynamic view routing. This may partly explain FlexWAM’s modest gains over strong baselines. Together, these results show that \textbf{FlexWAM learns to adapt view weights to the action required at each manipulation stage, with the largest gains on high-precision tasks.}


\subsection{Failure Analysis}

\begin{figure}[t]
    \centering
    \includegraphics[width=1\linewidth]{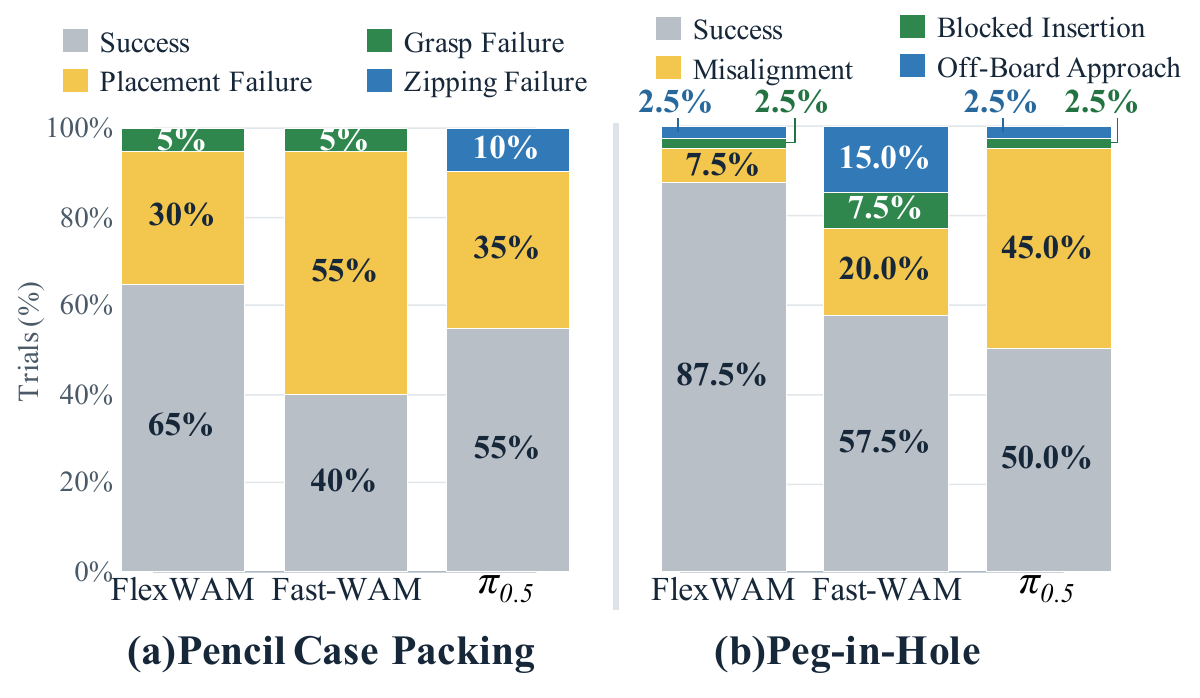}
    \vspace{-0.5em}
    \captionsetup{skip=2pt}
    \caption{Failure modes for (a) pencil case packing (20 trials) and (b) peg-in-hole insertion (40 trials per method across L-shaped and round pegs).
    FlexWAM uses the Wan backbone.}
    \label{fig:failure}
    \vspace{-1.5em}
\end{figure}

\begin{table*}[t]
    \centering
    \caption{Task success rates (\%) across attention schedules. Global Only ($k=1$), Alternating ($k=2$), and Sparse Cross-View ($k=4$) repeat zero, one, and three Local Within-View Attention layers, respectively,
    followed by one Global Cross-View Attention layer.
    Pens Placement averages placement success across the two pens.
    Pencil Case Overall requires successful packing and zipper closure.}
    \label{tab:attention_ablation}
    \footnotesize
    \setlength{\tabcolsep}{3pt}
    \renewcommand{\arraystretch}{1.1}
    \begin{tabular*}{\textwidth}{@{\extracolsep{\fill}}lccccccc@{}}
        \toprule
        \multirow{2}{*}{Attention Schedule}
        & \multirow{2}{*}{\shortstack{Cross-View\\Interval $k$}}
        & \multicolumn{3}{c}{\textbf{Wan / No Router}}
        & \multicolumn{2}{c}{\textbf{Wan / With Router}}
        & \textbf{LingBot / With Router} \\
        \cmidrule(lr){3-5}
        \cmidrule(lr){6-7}
        \cmidrule(lr){8-8}
        & 
        & \textbf{\textit{Pens Placement}}
        & \shortstack{\textbf{\textit{Pencil Case}}\\\textbf{\textit{Overall}}}
        & \shortstack{\textbf{\textit{Peg-in-Hole}}\\\textbf{\textit{(L)}}}
        & \textbf{\textit{Pens Placement}}
        & \shortstack{\textbf{\textit{Pencil Case}}\\\textbf{\textit{Overall}}}
        & \shortstack{\textbf{\textit{Cable}}\\\textbf{\textit{Insertion}}} \\
        \midrule
        Global Only
        & 1
        & 55.0\% & 40.0\% & 60.0\% & 77.5\% & \textbf{65.0\%} & \textbf{75.0\%} \\
        Alternating
        & 2
        & \textbf{70.0\%} & \textbf{45.0\%} & \textbf{70.0\%} & \textbf{80.0\%} & \textbf{65.0\%} & \textbf{75.0\%} \\
        Sparse Cross-View
        & 4
        & 57.5\% & \textbf{45.0\%} & 55.0\% & 47.5\% & 30.0\% & 55.0\% \\
        \bottomrule
    \end{tabular*}
    \vspace{-1em}
\end{table*}

\textbf{Failure analysis illustrates how counterfactually calibrated view routing supports manipulation by adapting visual contributions to changing action requirements (Q2).}
In pencil case packing, FlexWAM reduces the placement failures that dominate the baseline results (Fig.~\ref{fig:failure}(a)). The baselines' placement offsets indicate insufficient positioning precision at the narrow opening, whereas FlexWAM prioritizes the active arm's wrist view to emphasize the pen--opening relationship and improve placement accuracy (Fig.~\ref{fig:router}(c)). For peg insertion, misalignment accounts for the largest share of failures in both baselines, with Fast-WAM~\cite{yuan2026fast} additionally exhibiting off-board approaches and blocked insertions (Fig.~\ref{fig:failure}(b)). FlexWAM increases wrist-view weight during alignment and insertion, supporting fine positional adjustments based on the relative geometry of the peg and hole (Fig.~\ref{fig:router}(b)). For cable insertion, FlexWAM alternates emphasis between the side and wrist views to improve coarse positioning and fine alignment for precise insertion (Fig.~\ref{fig:router}(a)), while in conveyor sorting, it rapidly adjusts view weights to stabilize grasping and placement (Fig.~\ref{fig:router}(d)). These results suggest that counterfactually calibrated view routing helps FlexWAM use complementary observations to generate more reliable actions by allocating visual contributions according to the relative importance of different views for the current manipulation stage.

\subsection{Ablation Studies}

\begin{figure}[t]
    \centering
    \includegraphics[width=1\linewidth]{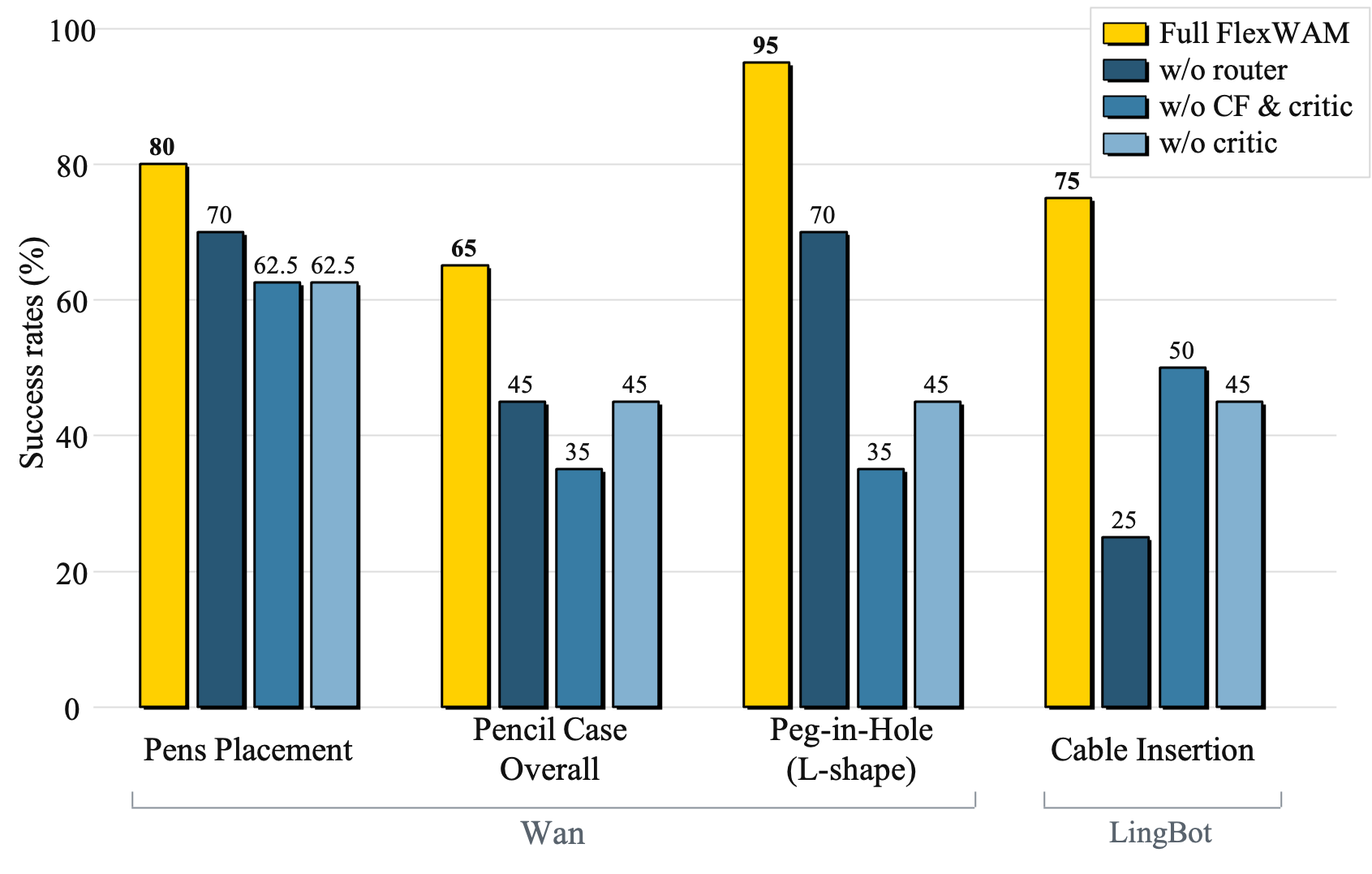}
    \captionsetup{skip=2pt}
    \caption{Ablation of view routing and counterfactual calibration. We compare the full FlexWAM model with three ablations: \textbf{Uniform Routing} (\textit{w/o router}), which removes the router and assigns uniform routing weights; \textbf{Action-Loss Routing} (\textit{w/o CF \& critic}), which updates the router using only action-loss gradients; and \textbf{Direct CF} (w/o critic), which trains the router jointly with action loss and KL calibration based directly on CF scores, using same KL warm-up parameters and duration.}
    \label{fig:router_ablation}
    \vspace{-1.5em}
    
\end{figure}

\textbf{Alternating Local Within-View Attention and Global Cross-View Attention achieves the best or tied-best task success across the evaluated settings (Q3)}. Its gains over Global Only are larger without routing and more modest with routing. As shown in Table~\ref{tab:attention_ablation}, its gain over Global Only on Pens Placement with Wan is 15.0 percentage points without routing and 2.5 points with routing. Without routing, alternating layers give the action expert access to both local detail preserved by Local Within-View Attention and complementary context integrated by Global Cross-View Attention. With routing, the action expert can prioritize a relevant branch that already incorporates cross-view context under either schedule, which may narrow the performance gap. Yet Sparse Cross-View performs substantially worse with routing, suggesting that effective view selection depends on sufficiently informative branch representations. We therefore adopt Alternating because \textbf{it preserves view-specific structure, maintains frequent cross-view exchange, and achieves the strongest results with and without routing.}

\textbf{The router enables adaptive weighting, CF provides action-grounded view credit, and the critic stabilizes router calibration (Q4).} As shown in Fig.~\ref{fig:router_ablation}, \textbf{Action-Loss Routing} improves Cable Insertion over uniform routing but lowers success on the other three metrics.
The learned weights stay near their uniform initialization. The action expert can fit demonstrations by adapting its own parameters, leaving the router little incentive to distinguish which views matter for each decision. \textbf{Critic-guided calibration improves success across all four metrics.} Both Direct CF and full FlexWAM train the router with action loss and KL calibration. \textbf{Direct CF} uses CF scores from individual evaluations, so its calibration target can vary with sampled action noise even for the same observation. In contrast, FlexWAM trains a critic on scores obtained from CF evaluations to estimate the expected view credit conditioned on current visual observations. The critic’s predictions provide calibration targets for the router, \textbf{mitigating fluctuations in supervision caused by action-noise sampling in individual CF evaluations.} Full FlexWAM outperforms Direct CF by 17.5–50.0 percentage points across all four metrics.

\section{Conclusion}
In this work, we presented FlexWAM to address the changing visual requirements of multi-camera robotic manipulation. FlexWAM learns complementary visual representations through view-specific dynamics modeling and cross-view interaction. Building on these representations, a critic trained on counterfactual view credit guides a flexible view router to adapt each camera's contribution to action generation. Real-world experiments on single-arm and dual-arm platforms demonstrate improved task success, particularly in manipulation requiring high precision, alongside adaptive view use during bimanual operation. Overall, FlexWAM establishes a framework for learning flexible visual routing for action generation, enabling world action models to translate complementary camera observations into more precise and reliable robotic behavior.


\section{Limitations and Future Work}
\label{sec:limitations}

\noindent\textbf{Scaling Visual Routing.}
Our evaluation covers five manipulation tasks across two robot platforms.
The effectiveness of FlexWAM during large-scale robot pretraining and its
transfer to broader task and embodiment distributions remain to be established.
Future work could investigate pretraining on more diverse robotic datasets
and systematically evaluate transfer to unseen tasks and embodiments.

\noindent\textbf{More Modality Routing.}
We currently explore adaptive routing among visual camera views.
Contact-rich manipulation may also benefit from tactile and force feedback,
particularly when visual evidence is insufficient.
Future work could extend counterfactual credit assignment and adaptive routing
to additional sensory modalities, enabling the policy to adjust its reliance
on vision, touch, and force as interaction demands change.

\noindent\textbf{Extending Beyond WAMs.}
We have so far evaluated flexible visual routing within world action models.
An interesting direction for future work is to extend this idea to
vision-language-action models, agentic robotic systems, and reinforcement
learning policies, exploring its broader applicability to robotic perception
and decision-making.

\section{Acknowledgement}
We would like to thank Ding Pan and Haoyue Zhang from Zhejiang University for valuable discussions. We also thank Jingqi Wang from Zhejiang University for assistance with experimental evaluation.
We are grateful to Alibaba Group and Endless AI for providing
computational resources to support this research.

{
    \small
    \bibliographystyle{ieeenat_fullname}
    \bibliography{ref}
}


\clearpage
\clearpage

\twocolumn[{%
    \centering
    {\Large\bfseries SUPPLEMENTARY MATERIALS FOR FlexWAM\par}
    \vspace{1em}
}]
\appendix
\setlength{\intextsep}{6pt}
\renewcommand{\thesection}{\Alph{section}}
\renewcommand{\thesubsection}{\thesection.\arabic{subsection}}
\setcounter{table}{0}
\setcounter{figure}{0}
\renewcommand{\thetable}{S\arabic{table}}
\renewcommand{\thefigure}{S\arabic{figure}}
\renewcommand{\theHtable}{appendix.\arabic{table}}
\renewcommand{\theHfigure}{appendix.\arabic{figure}}
\setcounter{equation}{0}
\renewcommand{\theequation}{S\arabic{equation}}
\renewcommand{\theHequation}{appendix.\arabic{equation}}

\section{Architecture and Training Details}
\label{sec:appendix_architecture}

\subsection{Model Configuration}
\label{sec:appendix_experts}

FlexWAM consists of a video expert and an action expert, jointly trained on multi-camera observations following the video--action framework of Fast-WAM~\cite{yuan2026fast}. We instantiate the video expert with Wan2.2-TI2V-5B~\cite{wan2025,wan2025wan22} or LingBot-Video-Dense (1.3B)~\cite{ma2026lingbotvideo,robbyant2026lingbotdense}. Images from each camera are resized to a resolution of $192 \times 224$ pixels and independently encoded using a shared frozen video VAE. The resulting visual tokens are augmented with camera embeddings and processed by alternating within-view and cross-view attention layers, inspired by VGGT~\cite{wang2025vggt} and detailed in Sec.~\ref{sec:view_modeling}. The action expert predicts a 32-step action chunk conditioned on current-frame visual features, language, and robot state. Table~\ref{tab:expert_configuration} summarizes the model configurations for the Wan and LingBot backbones.

\begin{table}[!ht]
    \centering
    \caption{Model configurations.}
    \label{tab:expert_configuration}
    \small
    \setlength{\tabcolsep}{4pt}
    \renewcommand{\arraystretch}{1.12}
    \begin{tabular}{@{}llcc@{}}
        \toprule
        Module & Parameter & Wan & LingBot \\
        \midrule
        \multirow{3}{*}{Video DiT} & Blocks & 30 & 24 \\
         & Hidden dimension & 3072 & 2048 \\
         & FFN dimension & 14336 & 6144 \\
        \midrule
        \multirow{3}{*}{Action DiT} & Blocks & 30 & 24 \\
         & Hidden dimension & 1024 & 2048 \\
         & FFN dimension & 4096 & 6144 \\
        \midrule
        \multirow{2}{*}{Both} & Heads & 24 & 16 \\
         & Head dimension & 128 & 128 \\ \midrule \multirow{2}{*}{Conditioning} & Language dimension & 4096 & 2560 \\ & State token dimension & 1024 & 2048 \\
        \bottomrule
    \end{tabular}
\end{table}

\subsection{Attention Visibility}
\label{sec:appendix_attention_visibility}

\textbf{Conditioning.} The current robot state is linearly projected into a single token prepended to the action sequence. This token can attend to language conditioning and current-frame visual tokens, but not to action or future visual tokens. Language conditioning is accessible to visual and action tokens but receives no information from either.

\noindent\textbf{Visual attention.} Both backbones use a first-frame causal mask following Fast-WAM~\cite{yuan2026fast}. The first latent frame corresponds to the clean current observation, and its visual tokens cannot attend to future visual tokens. Tokens in each future latent frame can attend to the current frame and all future frames. These temporal rules apply within the same camera in within-view layers and across all cameras in cross-view layers. Visual tokens do not attend to action or robot-state tokens.

\noindent\textbf{Action attention.} Action tokens attend bidirectionally within the entire 32-step chunk and access current-frame visual tokens from all cameras, together with language and the robot-state token; future visual tokens are masked out. Thus, future-frame information cannot reach action predictions.

\subsection{View Router and Counterfactual Critic}
\label{sec:appendix_router_critic}

\begin{figure}[!ht]
    \centering
    \includegraphics[width=0.7\columnwidth]{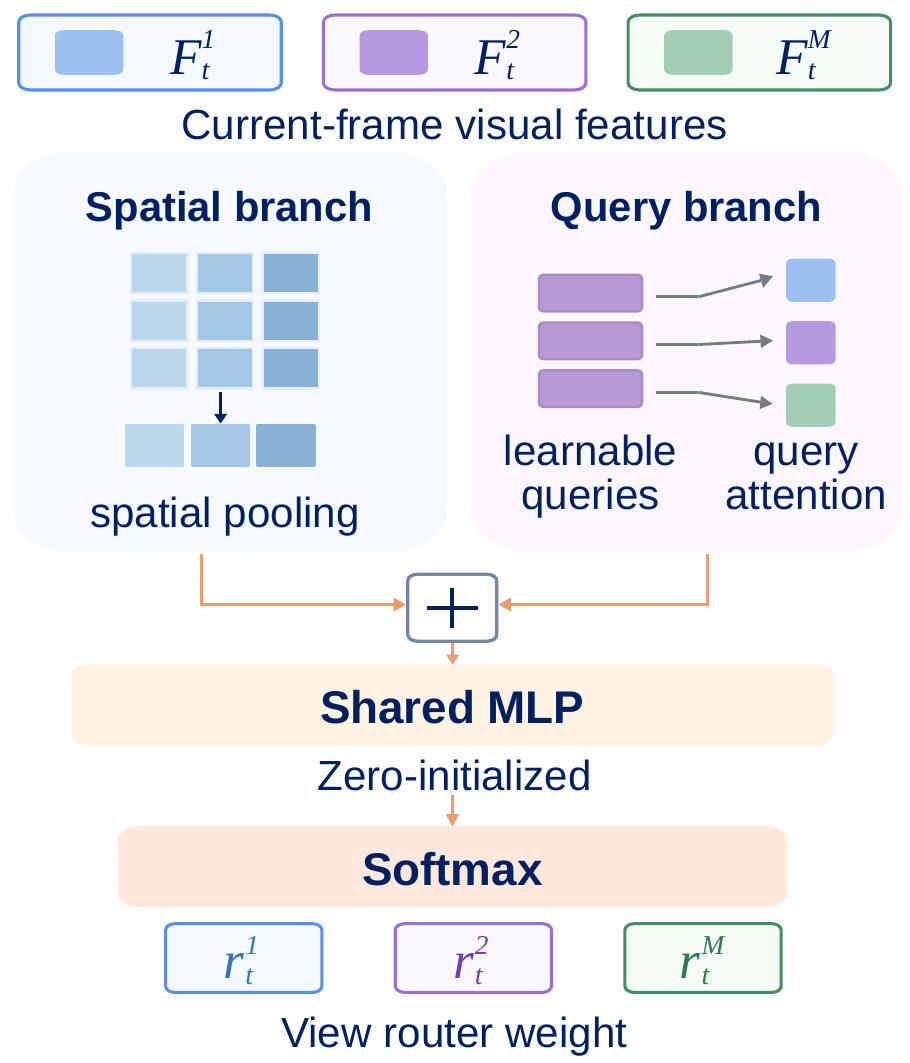}
    \caption{Network architecture shared by the view router and CF critic.}
    \label{fig:router_critic_architecture}
\end{figure}

\noindent\textbf{Network architecture.}
The view router and counterfactual (CF) critic share the same feature-processing and prediction architecture but have independent parameters and different output transformations. Both take current-frame features from Video DiT block 21 for Wan and 19 for LingBot. These features are detached to prevent router and critic supervision from updating the video expert through this input path. Layer normalization~\cite{ba2016layer}, linear projection, and GELU~\cite{hendrycks2016gelu} map per-view tokens to 512-dimensional features for the two aggregation branches. Both networks use zero-initialized weights and biases in the final prediction layer. Fig.~\ref{fig:router_critic_architecture} illustrates the router, including its output softmax.

\noindent\textbf{Feature aggregation.}
 As shown in Fig.~\ref{fig:router_critic_architecture}, two parallel branches combine coarse spatial layout with localized cues relevant to the manipulation stage. The spatial branch adaptively averages features onto a $4\times4$ grid, then flattens them for a two-layer MLP with GELU. The query branch uses eight-head attention~\cite{vaswani2017attention}, with four learned queries as $Q$, projected features with 2D sinusoidal positional embeddings as $K$, and projected features as $V$. Its outputs are concatenated and projected to 1024 dimensions with GELU. The two 1024-dimensional branch outputs form a concatenated per-view summary; the input projection and both branches are shared across views within each network. Summaries from all $M$ views ($M=3$ in our experiments) are concatenated for a two-layer joint prediction MLP with GELU, allowing each output to depend on all cameras. Table~\ref{tab:router_critic_configuration} lists the dimensions.

\par\smallskip\noindent
\begin{minipage}{\columnwidth}
    \captionsetup{type=table}
    \centering
    \caption{Router and critic configurations.}
    \label{tab:router_critic_configuration}
    \small
    \setlength{\tabcolsep}{1.5pt}
    \renewcommand{\arraystretch}{1.12}
    \begin{tabular}{lc}
        \toprule
        Configuration & Dimension \\
        \midrule
        Input features (Wan / LingBot) & 3072 / 2048 \\
        Spatial-branch MLP & $(4\times4\times512)\rightarrow2048\rightarrow1024$ \\
        Query projection & $(4\times512)\rightarrow1024$ \\
        Joint prediction MLP & $(M\times2048)\rightarrow1024\rightarrow M$ \\
        \bottomrule
    \end{tabular}
\end{minipage}
\par\medskip

\noindent\textbf{View routing.}
Softmax produces the router's view probabilities $r_t^m$, used as log-biases $\log(Mr_t^m)$ in the corresponding visual attention branches of the action expert. Zero initialization yields uniform routing and zero attention biases.

\noindent\textbf{Critic supervision.}
The critic outputs one unconstrained score per view without softmax and regresses normalized CF scores obtained from counterfactual evaluations. Its predictions are converted through a temperature-scaled softmax into detached soft targets for router calibration. Deployment requires only the router.

\subsection{Four-Stage Training and Deployment}
\label{sec:appendix_training}

The video and action experts are jointly optimized throughout all four stages, with stage boundaries specified in Table~\ref{tab:training_stages}. Equation~\eqref{eq:parameter_objectives} defines the objectives for the active parameter groups. CF evaluation and critic updates occur every five optimization steps in Stages~3 and~4; router KL calibration is additionally applied at those steps in Stage~4.

\begin{table}[!ht]
    \centering
    \caption{Four-stage training schedule.}
    \label{tab:training_stages}
    \small
    \setlength{\tabcolsep}{3pt}
    \renewcommand{\arraystretch}{1.12}
    \begin{tabular}{@{}ccll@{}}
        \toprule
        Stage & Training progress & Router & Critic \\
        \midrule
        1 & $0$--$10\%$ & Uniform  & Inactive \\
        2 & $10$--$30\%$ & Action loss & Inactive \\
        3 & $30$--$60\%$ & Action loss & CF regression \\
        4 & $60$--$100\%$ & Action loss + KL & CF regression \\
        \bottomrule
    \end{tabular}
\end{table}

\textbf{Stage 1: Expert warm-up (0--10\%).}
Camera gates are fixed to $1{:}1{:}1$, giving all three views equal weight. This stage stabilizes action learning and improves the video expert's visual representations before adaptive routing.

\textbf{Stage 2: Action-driven routing (10--30\%).}
The router is trained through action-loss gradients to learn observation-dependent view weights, while the critic remains inactive.

\textbf{Stage 3: Critic warm-up (30--60\%).}
The critic learns to predict normalized counterfactual (CF) scores by regression, while the router continues to learn from the action loss. CF scores measure the action-loss change caused by masking each view's visual branch in action attention. CF evaluation and critic updates occur every five optimization steps.

\textbf{Stage 4: Critic-guided calibration (60--100\%).}
Router training combines the action loss with KL calibration toward detached, critic-derived soft targets. CF evaluation and critic updates continue every five steps, with KL calibration applied at the same steps. The KL loss updates only the router. Throughout this stage, the KL coefficient increases linearly from $0.33$ to $0.5$, while the target temperature decreases linearly from $2$ to $1$.

\textbf{Deployment.}
Current-frame visual features and routing weights are computed once per observation and reused across all action-denoising steps. Future-video generation, CF evaluation, and the critic are omitted.

\section{Counterfactual Evaluation}
\label{sec:appendix_cf}

\subsection{Implementation of CF Evaluation}
\label{sec:appendix_cf_implementation}

Following the intervention in Eq.~\eqref{eq:view_intervention}, we compare action prediction losses before and after masking each camera's visual branch in action attention. For camera $m$, the loss increment is
\begin{equation}
\label{eq:appendix_cf_delta}
\Delta\ell_{\mathrm{act},t}^{m}
=
\ell_{\mathrm{act},t}(\mathbf w^{(-m)})
-
\ell_{\mathrm{act},t}(\mathbf w^{\mathrm{ref}}),
\end{equation}
where the two terms denote the action losses with camera $m$'s visual branch masked in action attention and with all visual branches retained, respectively. Following Eq.~\eqref{eq:counterfactual_normalization}, the signed loss increments are standardized across views within each sample to obtain normalized CF scores $z_t^m$, which are detached to provide stop-gradient supervision for the critic.

For CF evaluation, we set all camera gates to one in the reference branch to avoid reinforcing the router's existing view preferences. We implement attention masking by setting the corresponding visual-token logits to $-\infty$, while keeping the other gates at one. The joint softmax renormalizes over the remaining visible tokens. For these evaluations, we sample action noise as
\begin{equation}
\label{eq:appendix_cf_noise}
\sigma\sim\mathcal U(0.6,1.0),
\qquad
\boldsymbol\epsilon\sim\mathcal N(\mathbf0,\mathbf I).
\end{equation}
The reference and masked branches share $\sigma$ and $\boldsymbol\epsilon$, evaluating losses on identical noisy actions.

The reference and masked branches share cached visual keys and values, reducing the cost of CF target construction. Each branch requires a single action-expert forward pass. With three cameras, a complete CF evaluation therefore consists of one visual encoding and video-expert forward pass, followed by four action-expert forward passes.

\subsection{Comparison of CF Interventions}
\label{sec:appendix_cf_interventions}

Although cross-view attention exchanges information among cameras,
each branch remains anchored to its originating view's tokens and
spatial structure. We compare pixel and attention masking to examine
how intervention choice affects CF evaluation under this feature sharing.

Pixel masking removes visual content from all input frames of the
selected camera before encoding and recomputes visual features with
view weights unchanged. This intervention can propagate across views
and may shift the feature distribution received by the action expert.
Attention masking reuses the original visual keys and values and masks
the selected branch's logits in action attention.
It preserves the world model's representations while changing their
use by the action expert, directly matching the router's role.
Its CF scores therefore measure branch contribution given the remaining
features. Pixel masking reruns both the visual and action branches,
whereas attention masking reruns only the action branch.

We evaluate both interventions on conveyor sorting with the dual-arm PiPER using Wan and L-shaped peg insertion with the single-arm Flexiv using LingBot, covering both robot platforms and backbones. Table~\ref{tab:appendix_cf_interventions} reports their success rates, together with CF evaluation times measured under the same H200 hardware and evaluation configuration. Attention masking achieves comparable or higher success rates on both tasks and provides approximately a $2.19\times$ speedup in CF evaluation on conveyor sorting.

\begin{table}[!ht]
    \centering
    \caption{Success rates and CF evaluation time by intervention.}
    \label{tab:appendix_cf_interventions}
    \small
    \setlength{\tabcolsep}{2pt}
    \renewcommand{\arraystretch}{1.12}
    \begin{tabular*}{\columnwidth}{@{\extracolsep{\fill}}lccc@{}}
        \toprule
        CF intervention & \shortstack{Conveyor\\sorting} & \shortstack{L-shaped\\insertion} & Time per CF \\
        \midrule
        Pixel masking & $85.0\%$ & $85.0\%$ & $\approx494$ ms \\
        \textbf{Attention masking(ours)} & $86.0\%$ & $95.0\%$ & $\approx225$ ms \\
        \bottomrule
    \end{tabular*}
\end{table}

\subsection{Noise Range and Target Stability}
\label{sec:appendix_cf_stability}

The noise range used in CF evaluation affects how much action information
remains available to the predictor. At low noise levels, noisy actions
retain substantial ground-truth information, allowing the model to rely
less on visual conditioning. Consequently, loss changes from masking
a camera's visual branch may not fully reflect visual contributions.
We restrict noise to $[0.6,1.0]$ to reduce these direct action cues and
encourage greater reliance on visual conditioning during CF evaluation.

We assess target stability on pencil case packing and cable insertion.
With models and inputs fixed, we repeatedly sample action noise and
measure the variance of each camera's $z_t^m$; lower variance indicates
less noise-induced variation and more stable critic supervision.
We use $\mathcal U(0,1)$ as the baseline, matching the action-training
noise distribution with shift $1$.
Table~\ref{tab:appendix_cf_stability} aggregates results across cameras
and input samples. Sampling from $\mathcal U(0.6,1)$ reduces variance
by $77.0\%$ and $94.3\%$ on the two tasks, respectively.
We therefore adopt $[0.6,1.0]$ to improve target stability while retaining
broad coverage of high-noise conditions.

\begin{table}[!ht]
    \centering
    \caption{Variance of $z_t^m$ under different noise sampling settings.}
    \label{tab:appendix_cf_stability}
    \small
    \setlength{\tabcolsep}{5pt}
    \renewcommand{\arraystretch}{1.12}
    \begin{tabular}{@{}lcc@{}}
        \toprule
        Task & $\mathcal U(0,1)$ & $\mathcal U(0.6,1)$ \\
        \midrule
        Pencil case packing & 0.15782 & 0.03627 \\
        Cable insertion & 0.23648 & 0.01358 \\
        \bottomrule
    \end{tabular}
\end{table}

\section{Ablation Studies and Analysis}
\label{sec:appendix_critic_calibration}
\subsection{Ablation Settings}
\label{sec:appendix_ablation_settings}

Table~\ref{tab:appendix_ablation_stages} summarizes the training-stage allocations for the ablation variants and full FlexWAM, with stage definitions provided in Table~\ref{tab:training_stages}. Uniform Routing uses uniform camera weights throughout training. Action-Loss Routing learns routing from the action loss alone. Early Direct CF adds KL calibration using CF scores immediately after Stage 1, whereas Direct CF postpones calibration until Stage 4. Full FlexWAM trains the critic in Stage 3 and introduces critic-guided calibration in Stage 4.

\par\smallskip\noindent
\begin{minipage}{\columnwidth}
    \captionsetup{type=table}
    \centering
    \caption{Training-stage allocations across settings. Percentages indicate fractions of total training, and stage labels refer to the optimization regimes in Table~\ref{tab:training_stages}. For Early Direct CF and Direct CF, Stage~4 uses direct CF targets for router calibration without a critic. Direct CF uses the same KL warm-up parameters and duration as full FlexWAM.}
    \label{tab:appendix_ablation_stages}
    \small
    \setlength{\tabcolsep}{3pt}
    \renewcommand{\arraystretch}{1.12}
    \begin{tabular}{@{}lcccc@{}}
        \toprule
        Setting & Stage 1 & Stage 2 & Stage 3 & Stage 4 \\
        \midrule
        Uniform Routing & $100\%$ & --- & --- & --- \\
        Action-Loss Routing & $10\%$ & $90\%$ & --- & --- \\
        Early Direct CF & $10\%$ & --- & --- & $90\%$ \\ Direct CF & $10\%$ & $50\%$ & --- & $40\%$ \\ \textbf{Full FlexWAM} & $10\%$ & $20\%$ & $30\%$ & $40\%$ \\
        \bottomrule
    \end{tabular}
\end{minipage}
\par\medskip

\subsection{Limitations of Direct CF Guidance}
\label{sec:appendix_direct_cf_limitations}

We compare direct CF guidance with critic-guided calibration on L-shaped peg insertion and Ethernet cable insertion with the single-arm Flexiv and pencil case packing with the dual-arm PiPER, covering both robot platforms. All experiments use Wan for L-shaped insertion and pencil case packing and LingBot for cable insertion, following Fig.~\ref{fig:router_ablation}. Table~\ref{tab:appendix_calibration_success} shows that direct CF guidance underperforms full FlexWAM across tasks whether calibration begins early or is delayed to match the full method's schedule.

\par\smallskip\noindent
\begin{minipage}{\columnwidth}
    \captionsetup{type=table}
    \centering
    \caption{Real-world success rates by calibration setting. }
    \label{tab:appendix_calibration_success}
    \small
    \setlength{\tabcolsep}{1.5pt}
    \renewcommand{\arraystretch}{1.12}
    \begin{tabular*}{\columnwidth}{@{\extracolsep{\fill}}lcccc@{}}
        \toprule
        \raisebox{0.5\baselineskip}{Method} & \shortstack{L-shaped\\Insertion} & \shortstack{Cable\\Insertion} & \shortstack{Pens\\Placement} & \shortstack{Pencil Case\\Overall} \\
        \midrule
        Early Direct CF & $55.0\%$ & $45.0\%$ & $67.5\%$ & $45.0\%$ \\
        Direct CF & $45.0\%$ & $45.0\%$ & $62.5\%$ & $45.0\%$ \\
        \textbf{Full FlexWAM} & $95.0\%$ & $75.0\%$ & $80.0\%$ & $65.0\%$ \\
        \bottomrule
    \end{tabular*}
\end{minipage}
\par\medskip

To understand this performance gap, we examine how action noise affects direct CF supervision. Direct CF guidance converts camera contribution scores into calibration targets that can vary with action noise even for the same observation. Resampling action noise changes the noisy action input and can alter the loss increase caused by camera masking, leading to different camera-weight targets.

Table~\ref{tab:appendix_target_variation} reports the mean pairwise $L_1$ distance between calibration targets under action-noise resampling, with the model and observation fixed and noise levels drawn from $\mathcal U(0.6,1.0)$. CF targets vary under action-noise resampling across tasks. The critic yields zero target variation because its predictions depend only on fixed current-frame visual features and are invariant to action-noise resampling.

\par\smallskip\noindent
\begin{minipage}{\columnwidth}
    \captionsetup{type=table}
    \centering
    \caption{Target variation under action-noise resampling.}
    \label{tab:appendix_target_variation}
    \small
    \setlength{\tabcolsep}{1.5pt}
    \renewcommand{\arraystretch}{1.12}
    \begin{tabular*}{\columnwidth}{@{\extracolsep{\fill}}lccc@{}}
        \toprule
        Target & L-shaped insertion & Cable insertion & Pencil case packing \\
        \midrule
        CF & 0.04778 & 0.04853 & 0.06445 \\
        Critic & 0 & 0 & 0 \\
        \bottomrule
    \end{tabular*}
\end{minipage}
\par\medskip

We further examine whether the router can follow its supervision during training. Table~\ref{tab:appendix_calibration_kl} reports the mean KL calibration loss, which measures the discrepancy between router outputs and their calibration targets. Direct CF variants have larger mean KL than Full FlexWAM across all three tasks. Direct CF has $7.35\times$, $5.02\times$, and $4.82\times$ the mean KL of Full FlexWAM on L-shaped insertion, cable insertion, and pencil case packing, respectively. Together with the noise-resampling results, this supports the critic's role in stabilizing router calibration by providing consistent supervision that the router can more closely follow.

\par\smallskip\noindent
\begin{minipage}{\columnwidth}
    \captionsetup{type=table}
    \centering
    \caption{Mean router-calibration KL.}
    \label{tab:appendix_calibration_kl}
    \small
    \setlength{\tabcolsep}{1.5pt}
    \renewcommand{\arraystretch}{1.12}
    \begin{tabular*}{\columnwidth}{@{\extracolsep{\fill}}lccc@{}}
        \toprule
        Task & Early Direct CF & Direct CF & \textbf{Full FlexWAM} \\
        \midrule
        L-shaped insertion & 0.02158 & 0.01948 & 0.00265 \\
        Cable insertion & 0.05827 & 0.07153 & 0.01425 \\
        Pencil case packing & 0.02415 & 0.03044 & 0.00631 \\
        \bottomrule
    \end{tabular*}
\end{minipage}
\par\medskip

\subsection{Resolution Effects in Image-Concat WAMs} \label{sec:appendix_resolution_allocation} \par We evaluate how the choice of view receiving the highest resolution affects Fast-WAM~\cite{yuan2026fast} on L-shaped peg insertion using the three settings in Table~\ref{tab:appendix_resolution_allocation}. The wrist setting achieves a $70\%$ success rate, outperforming the global-view and side-view settings by 10 and 60 percentage points, demonstrating Fast-WAM's sensitivity to which view receives the highest resolution. The improvement from increasing wrist-image resolution suggests that fine-grained wrist-view information benefits precise insertion, consistent with the greater wrist-view weight learned by FlexWAM on this task. Rather than manually assigning higher resolution to a selected camera, FlexWAM independently encodes all views at a common resolution and learns their relative contributions, allowing it to emphasize task-relevant visual features as the manipulation stage changes. \par\smallskip\noindent \begin{minipage}{\columnwidth} \captionsetup{type=table} \centering \caption{Fast-WAM success rates at different resolutions.} \label{tab:appendix_resolution_allocation} \small \setlength{\tabcolsep}{3pt} \renewcommand{\arraystretch}{1.12} \begin{tabular}{@{}cccc@{}} \toprule Global resolution & Side resolution & Wrist resolution & Success rate \\ \midrule $\mathbf{256\times320}$ & $128\times160$ & $128\times160$ & $60.0\%$ \\ $128\times160$ & $\mathbf{256\times320}$ & $128\times160$ & $10.0\%$ \\ $128\times160$ & $128\times160$ & $\mathbf{256\times320}$ & $70.0\%$ \\ \bottomrule \end{tabular} \end{minipage} \par\medskip 

\section{Additional Experimental Details}
\label{sec:appendix_experimental_details}

\subsection{Evaluation Setups and Protocol}
\label{sec:appendix_evaluation_setups}

\begin{figure*}[!t]\centering\includegraphics[width=\textwidth]{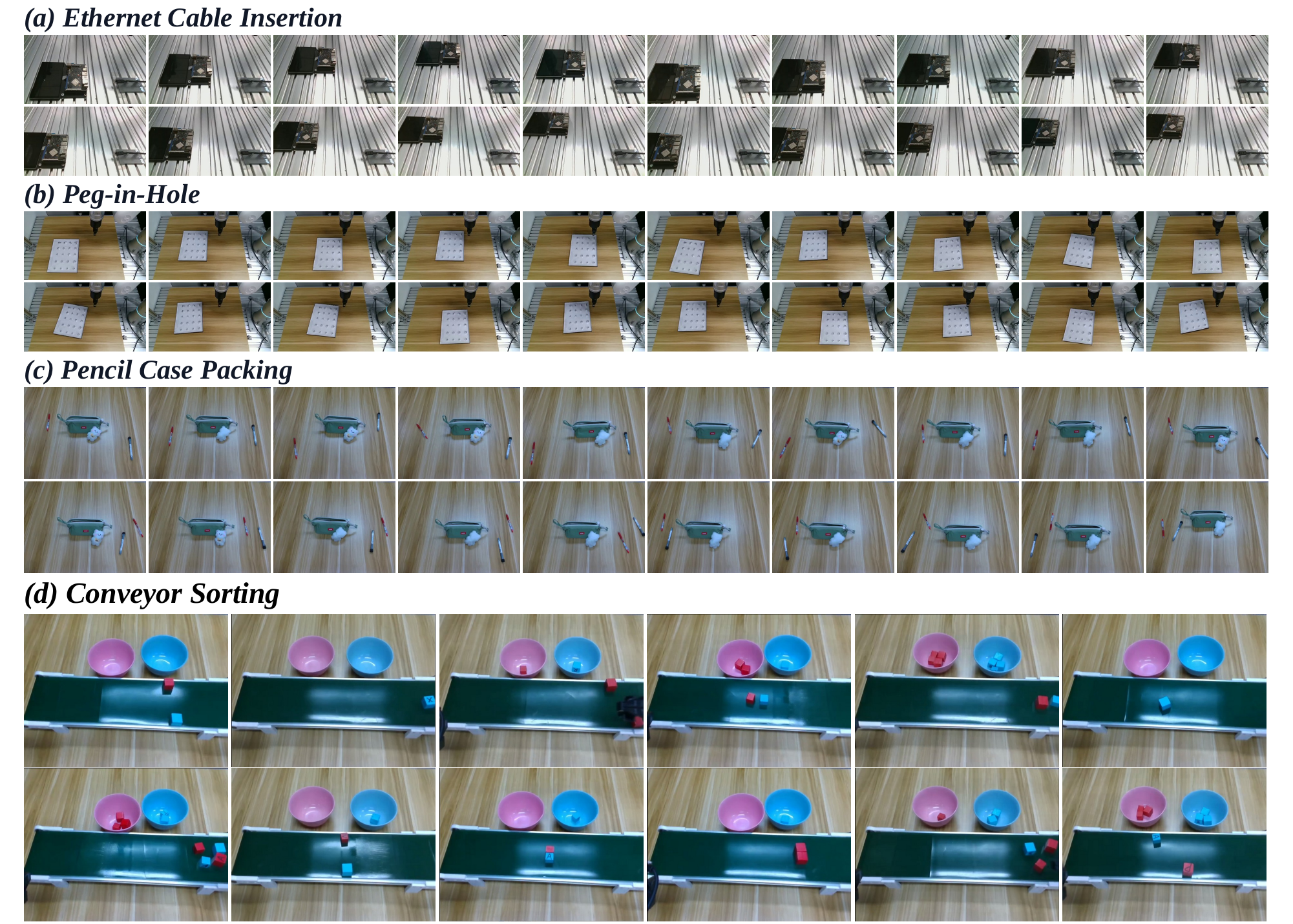}
\caption{Real-world evaluation setups.}
\label{fig:appendix_evaluation_setups}
\vspace{-1em}
\end{figure*}

Figure~\ref{fig:appendix_evaluation_setups} shows the object arrangements used for real-world evaluation. We conduct 20 trials each for Ethernet cable insertion, L-shaped peg insertion, round peg insertion, and pencil case packing. Both peg-in-hole tasks use the $0.02$~mm clearance setting from ManipulationNet~\cite{chen2026manipulationnet}. Insertion tasks are evaluated by successful insertion, while pencil case packing reports both pen-placement success and overall task completion. For conveyor sorting, we present 100 blocks in randomized combinations of the 12 illustrated arrangements and evaluate success by the proportion of blocks correctly sorted. All methods and ablations use the same randomized initial positions for the four preceding tasks and follow the same block-placement protocol for conveyor sorting.

\subsection{Baselines}
\label{sec:appendix_baselines}

We compare FlexWAM against three representative World Action Models (WAMs) and one Vision-Language-Action (VLA) model:

\begin{itemize}
    \item \textbf{Fast-WAM}~\cite{yuan2026fast}, a WAM that jointly trains video and action experts but predicts actions from current observations without future-video generation at inference.

    \item \textbf{OpenWAM}~\cite{wang2026openwam}, a WAM pretrained on large-scale egocentric human and robot data, coupling world modeling and action learning through dedicated action capacity and joint video--action denoising.

    \item \textbf{LaWAM}~\cite{chen2026lawam}, a latent WAM that conditions action generation on predicted latent visual subgoals, providing future context without reconstructing future video.

    \item \textbf{$\pi_{0.5}$}~\cite{pi05}, a VLA model co-trained on heterogeneous robotic and non-robotic data for generalizable language-conditioned manipulation.
\end{itemize}

\noindent\textbf{Action representations.}
OpenWAM and LaWAM use absolute  actions during pretraining; we retain this parameterization during fine-tuning and evaluation. Fast-WAM, $\pi_{0.5}$, and both FlexWAM variants use delta actions.

\subsection{Computational Efficiency}
\label{sec:appendix_efficiency}

We measure computational costs on an H200, generating a 32-step action chunk with 10 denoising steps. The image-concatenation baseline combines a $256\times320$ global view and two $128\times160$ wrist views into a $384\times320$ canvas and encodes it with the backbone's VAE. Independent encoding instead encodes each of the three $192\times224$ views separately with the same VAE. The Wan2.2~\cite{wan2025wan22} and LingBot~\cite{robbyant2026lingbotdense} VAEs spatially downsample by factors of $16$ and $8$, respectively, both followed by $2\times2$ spatial patchification.

Table~\ref{tab:appendix_inference_cost} compares inference costs using the same backbone and weights, with Tokens denoting the visual token count for the current observation. Image concatenation and Independent disable the router, whereas FlexWAM uses independent encoding with adaptive routing. Independent encoding increases visual tokens by only $5\%$, with CUDA Graph latency increasing by $10.0\%$ on Wan and $0.8\%$ on LingBot relative to image concatenation. FlexWAM adds a further 0.98 ms and 1.26 ms, respectively, indicating modest additional inference cost. CUDA Graph replay~\cite{gray2019cudagraphs} executes the captured action-denoising steps, accelerating FlexWAM inference by $2.78\times$ on Wan and $2.52\times$ on LingBot compared with eager execution.

\begin{table}[!ht]
    \centering
    \caption{Visual token counts and inference latency (ms).}
    \label{tab:appendix_inference_cost}
    \small
    \setlength{\tabcolsep}{3pt}
    \renewcommand{\arraystretch}{1.15}
    \begin{tabular}{@{}clrrr@{}}
        \toprule
        Backbone & Configuration & Tokens & Eager & CUDA Graph \\
        \midrule
        & Image concatenation & 120 & 343.77 & 118.13 \\
            Wan & Independent & 126 & 358.58 & 129.99 \\
            & \textbf{FlexWAM} & 126 & 364.43 & 130.97 \\
        \midrule
        & Image concatenation & 480 & 248.52 & 97.14 \\
                LingBot & Independent & 504 & 250.64 & 97.94 \\
                & \textbf{FlexWAM} & 504 & 250.21 & 99.20 \\
        \bottomrule
    \end{tabular}
\end{table}

\begin{table}[!ht]
    \centering
    \caption{Mean training-step time.}
    \label{tab:appendix_training_cost}
    \small
    \setlength{\tabcolsep}{5pt}
    \renewcommand{\arraystretch}{1.12}
    \begin{tabular}{@{}lcc@{}}
        \toprule
        Backbone & w/o router (ms/step) & \textbf{FlexWAM (ms/step)} \\
        \midrule
        Wan & 421.83 & 459.37 \\
        LingBot & 280.31 & 307.24 \\
        \bottomrule
    \end{tabular}
\end{table}

\begin{figure*}[!t]
    \centering
    \includegraphics[width=\textwidth]{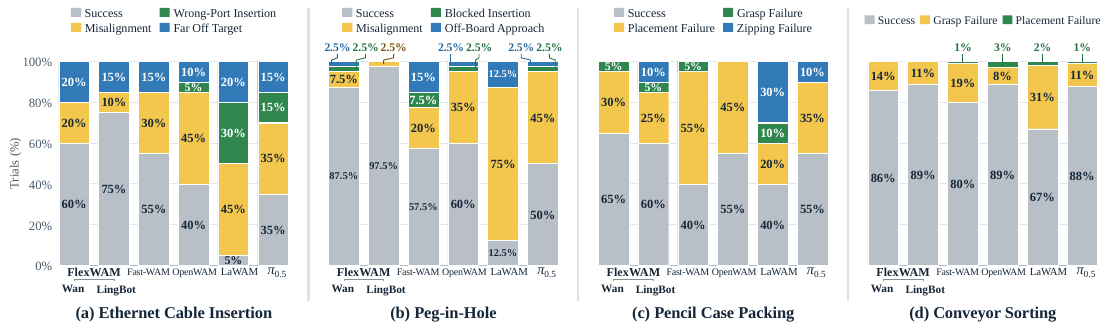}
    \caption{Real-world success and failure rates by task and method.}
    \label{fig:appendix_full_failure_analysis}
\end{figure*}

Table~\ref{tab:appendix_training_cost} compares training costs against a uniform-routing baseline without CF evaluation or critic updates. We estimate FlexWAM's average training-step time by weighting each stage's measured mean by its proportion of training, with CF evaluation performed every five steps in Stages 3 and 4. This yields an estimated training overhead of $8.9\%$ on Wan and $9.6\%$ on LingBot.

\subsection{Full Failure Analysis}
\label{sec:appendix_failure_analysis}
\begin{figure*}[!t]\centering\includegraphics[width=0.85\textwidth]{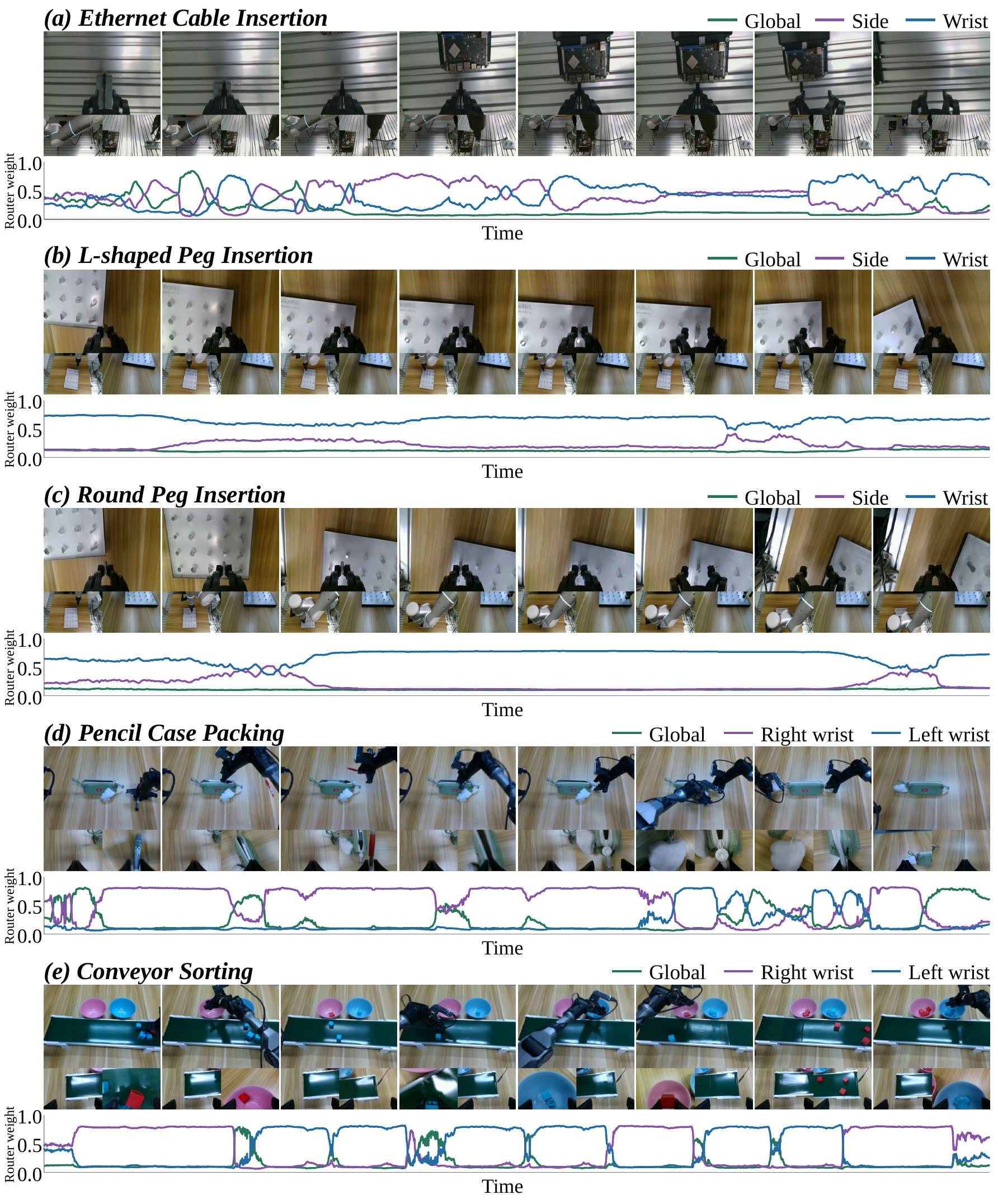}\caption{Router weights over complete task trajectories with illustrative task images.}\label{fig:appendix_router_dynamics}\end{figure*}

Figure~\ref{fig:appendix_full_failure_analysis} extends the failure analysis in Fig.~\ref{fig:failure} to all methods and both FlexWAM backbones, interpreted alongside the routing trajectories in Fig.~\ref{fig:appendix_router_dynamics}. Panel~(b) pools 40 trials per method across L-shaped and round pegs. For pencil case packing, trials with failed zipping are classified as zipping failures regardless of earlier grasp or placement failures. The grasp- and placement-failure categories include only trials with successful zipping.

\noindent\textbf{Ethernet cable insertion.}
Misalignment occurs in $30$--$45\%$ of baseline trials, versus $20\%$ for FlexWAM (Wan) and $10\%$ for FlexWAM (LingBot). Neither variant exhibits wrong-port insertion, although far-off-target failures persist at $20\%$ and $15\%$, respectively. Routing shifts from side-view emphasis during transport to wrist-view emphasis during alignment and insertion (Fig.~\ref{fig:appendix_router_dynamics}(a)), consistent with complementary roles in coarse positioning and local connector--port alignment.

\noindent\textbf{Peg-in-hole insertion.}
Both FlexWAM backbones achieve $95\%$ success on L-shaped insertion; round-insertion success reaches $80\%$ with Wan and $100\%$ with LingBot. Pooled misalignment rates decrease from $20$--$75\%$ across baselines to $7.5\%$ and $2.5\%$, respectively. Wan also retains $2.5\%$ blocked insertions and $2.5\%$ off-board approaches. Across both geometries, the router prioritizes the wrist view during fine alignment and insertion, consistent with an emphasis on local peg--hole geometry. Transient increases in side-view weight during approach and withdrawal provide complementary visual context (Fig.~\ref{fig:appendix_router_dynamics}(b,c)).

\noindent\textbf{Pencil case packing.}
Placement remains FlexWAM's dominant failure mode: $30\%$ with Wan and $25\%$ with LingBot, versus $55\%$ for Fast-WAM and $35\%$ for $\pi_{0.5}$. Active-wrist emphasis during grasping and placement is consistent with localizing the pen relative to the narrow opening, while global-view weight rises during approach and repositioning (Fig.~\ref{fig:appendix_router_dynamics}(d)). Zipping failures are absent with Wan but reach $10\%$ with LingBot, highlighting that placement gains do not ensure successful closure.

\noindent\textbf{Conveyor sorting.}
FlexWAM achieves $86\%$ success with Wan and $89\%$ with LingBot, improving over Fast-WAM's $80\%$ by 6 and 9 percentage points, respectively. The LingBot variant matches the strongest baseline, OpenWAM, at $89\%$. Placement failures are absent, versus $1$--$3\%$ across baselines; all remaining failures involve grasping ($14\%$ and $11\%$, respectively). Alternating active-wrist emphasis, with brief global-view increases between operations (Fig.~\ref{fig:appendix_router_dynamics}(e)), is consistent with reliable depositing and repositioning, while grasping moving blocks remains the principal limitation.

\makeatletter\setlength{\@fptop}{0pt}\setlength{\@dblfptop}{0pt}\makeatother

\subsection{Camera Weights over Time}
\label{sec:appendix_router_dynamics}

Figure~\ref{fig:appendix_router_dynamics} shows router weights over complete demonstration trajectories from five tasks. In pencil case packing, the global-view weight rises during approach, pen transfer toward the case, and withdrawal after task completion, while the active arm's wrist view dominates pen grasping and placement. Conveyor sorting shifts weight between the wrist views as the arms alternate in grasping and depositing blocks, with brief increases in global-view weight between these operations to provide workspace context for arm repositioning. In Ethernet cable insertion, wrist-view weight rises during grasping, while the global and side views gain weight during transport. The wrist view dominates fine alignment and insertion. Both peg-in-hole tasks similarly favor the wrist view during fine alignment and insertion, with transient increases in side-view weight during approach and withdrawal.

\end{document}